\documentclass{bmvc2k}
\usepackage{amsfonts}
\usepackage{amsmath}
\usepackage{comment}
\usepackage[algo2e,ruled,vlined,linesnumbered]{algorithm2e}
\usepackage{algorithmic} 
\usepackage[ruled,vlined,linesnumbered]{algorithm2e}

\title{Steady-Forcing: Balancing Spatial Persistence and Motion Continuity in Long-Horizon Nature Video Diffusion}

\addauthor{Matiur Rahman Minar}{minar@sogang.ac.kr}{1}
\addauthor{Seunghun Oh}{gnsgus190@sogang.ac.kr}{2}
\addauthor{Ganghyeon Jeong}{jugahy1205@sogang.ac.kr}{2}
\addauthor{Unsang Park}{unsangpark@sogang.ac.kr}{1,2}

\addinstitution{
 Department of Computer Science and Engineering\\
 Sogang University\\
 Seoul, Korea
}

\addinstitution{
 Department of Artifical Intelligence\\
 Sogang University\\
 Seoul, Korea
}

\runninghead{Minar et al.}{Steady-Forcing}

\def\eg{\emph{e.g}\bmvaOneDot}
\def\Eg{\emph{E.g}\bmvaOneDot}
\def\etal{\emph{et al}\bmvaOneDot}

\begin{document}

\maketitle

\begin{abstract}
Autoregressive video diffusion models enable streaming generation but often degrade over long rollouts: static scene layouts drift, while mechanisms that improve spatial stability tend to suppress motion, causing natural flows such as water, fire, or smoke to stagnate. We study this stability--motion trade-off in fixed-camera long-horizon nature video generation, where the two failure modes can be more clearly separated than in moving-camera settings. We propose Steady-Forcing, a memory and training framework combining a persistent visual anchor (V-Sink), an exponential moving-average motion memory (EMA-Sink), block-relative temporal encoding, periodic cache purification, and distillation from a Wan2.1-14B teacher with motion-rewarded priors under task-focused configurations. Together, these components are designed to preserve background identity while sustaining visually plausible fluid dynamics over multi-minute autoregressive rollouts. Evaluations across seven baselines show that Steady-Forcing improves long-horizon background consistency and imaging quality, while a blind user study indicates stronger perceived stability and motion continuity. The benchmark evaluation further suggest that generic VBench aggregate scores under-penalize fixed-camera artifacts as well as rewarding drift-induced optical flow as Dynamic Degree while not directly penalizing texture hardening or flow stagnation -- motivating future task-specific benchmarks for static-camera nature-flow evaluation. Project page: \url{https://minar09.github.io/steadyforcing/}
\end{abstract}


\begin{figure*}[t]
\centering
\setlength{\tabcolsep}{1pt}
\begin{tabular}{@{}c@{\hspace{2pt}}cccc@{}}
& \scriptsize $t=0\text{s}$ & \scriptsize $t=20\text{s}$ & \scriptsize $t=40\text{s}$ & \scriptsize $t=60\text{s}$ \\

\rotatebox{90}{\hspace{-2em}\small Sea} &
\bmvaHangBox{\includegraphics[width=0.2\linewidth]{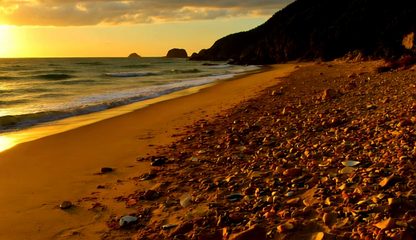}} &
\bmvaHangBox{\includegraphics[width=0.2\linewidth]{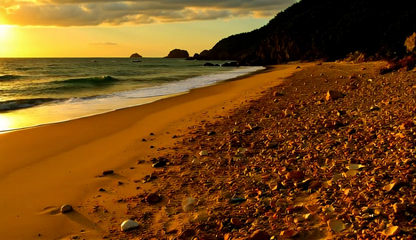}} &
\bmvaHangBox{\includegraphics[width=0.2\linewidth]{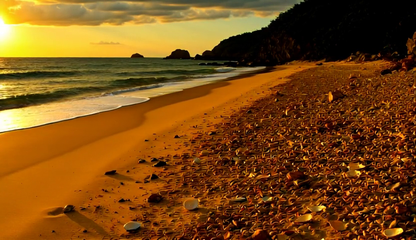}} &
\bmvaHangBox{\includegraphics[width=0.2\linewidth]{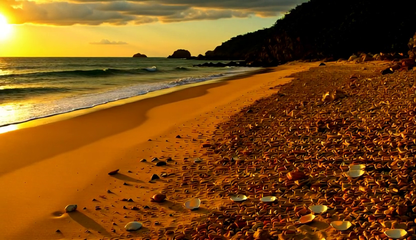}} \\

\vspace{-5mm} \\

\rotatebox{90}{\hspace{-2em}\small Snow} &
\bmvaHangBox{\includegraphics[width=0.2\linewidth]{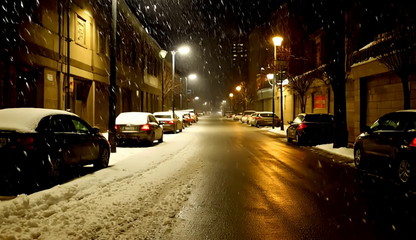}} &
\bmvaHangBox{\includegraphics[width=0.2\linewidth]{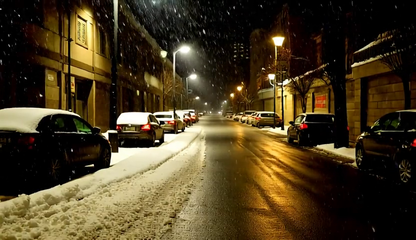}} &
\bmvaHangBox{\includegraphics[width=0.2\linewidth]{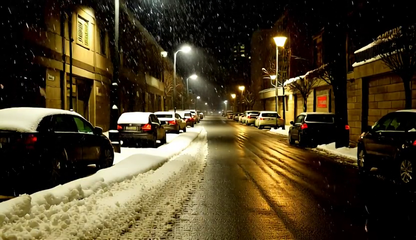}} &
\bmvaHangBox{\includegraphics[width=0.2\linewidth]{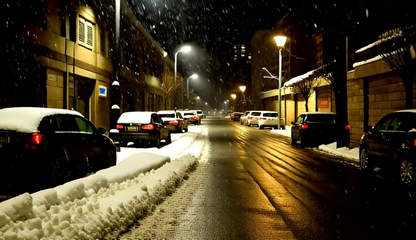}} \\

\vspace{-5mm} \\

\rotatebox{90}{\hspace{-2em}\small River} &
\bmvaHangBox{\includegraphics[width=0.2\linewidth]{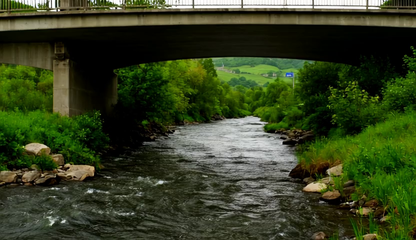}} &
\bmvaHangBox{\includegraphics[width=0.2\linewidth]{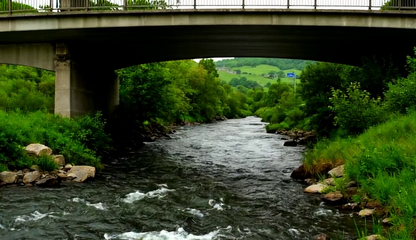}} &
\bmvaHangBox{\includegraphics[width=0.2\linewidth]{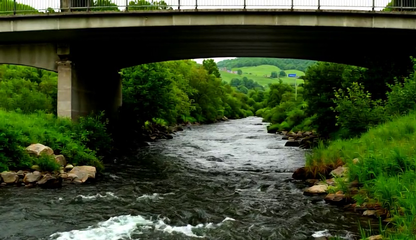}} &
\bmvaHangBox{\includegraphics[width=0.2\linewidth]{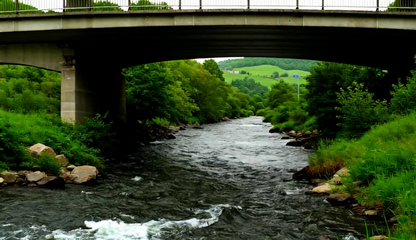}} \\

\vspace{-5mm} \\

\rotatebox{90}{\hspace{-2em}\small Storm} &
\bmvaHangBox{\includegraphics[width=0.2\linewidth]{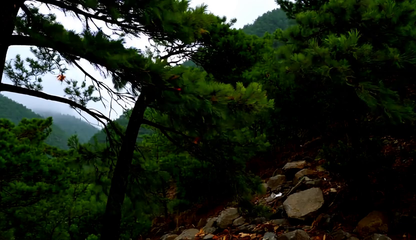}} &
\bmvaHangBox{\includegraphics[width=0.2\linewidth]{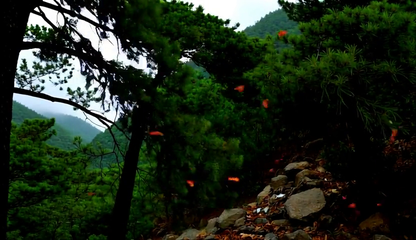}} &
\bmvaHangBox{\includegraphics[width=0.2\linewidth]{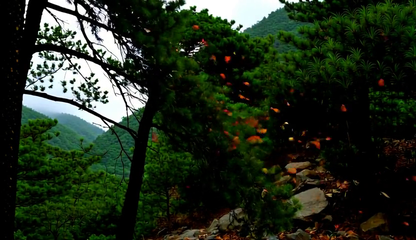}} &
\bmvaHangBox{\includegraphics[width=0.2\linewidth]{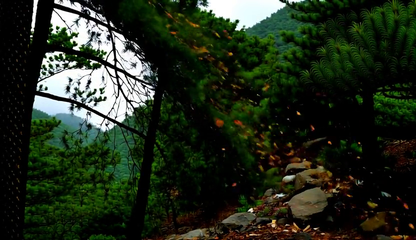}} \\

\end{tabular}
\caption{We propose Steady-Forcing to tackle the video generation task of long-horizon, static-scene, continuous-flow nature streams. The unified dual-sink and periodic purification mechanisms are designed to preserve spatial identity while maintaining visually plausible fluid motion over long durations.}
\label{fig:top_results}
\end{figure*}


\section{Introduction}
\label{sec:intro}

The scaling of Diffusion Transformers (DiTs), from CogVideoX's expert transformer architecture~\cite{yang2025cogvideox} to Wan2.2's Mixture-of-Experts formulation~\cite{wan2025}, has enabled photorealistic video synthesis at unprecedented quality. Yet most models remain constrained to short 5--10 second clips due to their reliance on full bidirectional temporal context. Recent research has therefore shifted toward \textbf{autoregressive (AR) video generation}~\cite{xiong2024autoregressive, ge2022long, hong2022cogvideo, kondratyuk2023videopoet, yan2021videogpt, yu2024language}, which factorizes the video distribution into a sequence of conditionals to enable low-latency streaming inference without future-frame dependency.

However, AR video DiTs encounter two compounding failure modes when extending generation horizons. First, \textbf{exposure bias}~\cite{huang2025selfforcing, bengio2015scheduled} causes prediction errors to accumulate across self-rollout steps,
progressively inducing background drift~\cite{camo_Cheong_2025_BMVC} and identity collapse~\cite{yin2025causvid}. Second, standard 3D Rotary Positional Embeddings (3D-RoPE)~\cite{su2024ropeformer} exhibit limited extrapolation beyond the temporal horizons seen during training, leading to degraded temporal attention and reduced stability during long autoregressive rollouts. While methods such as Infinite-Forcing~\cite{infinite-forcing} and Rolling Forcing~\cite{liu2025rolling} use attention sinks to mitigate spatial drift, they often suppress motion dynamics in the process. In nature scenes, this manifests as \textbf{motion stagnation}: fluid regions gradually lose temporal momentum, causing flowing water, fire, or clouds to converge toward static, frozen textures over extended rollouts.

\begin{figure*}[t]
\centering
\setlength{\tabcolsep}{3pt} 
\begin{tabular}{ccc}
\scriptsize $t=0\text{s}$ & \scriptsize $t=30\text{s}$ & \scriptsize $t=60\text{s}$ \\ [0em]

\bmvaHangBox{\includegraphics[width=0.2\linewidth]{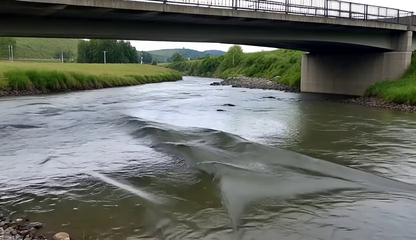}} &
\bmvaHangBox{\includegraphics[width=0.2\linewidth]{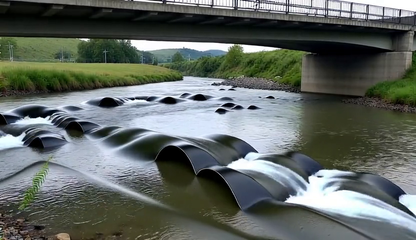}} &
\bmvaHangBox{\includegraphics[width=0.2\linewidth]{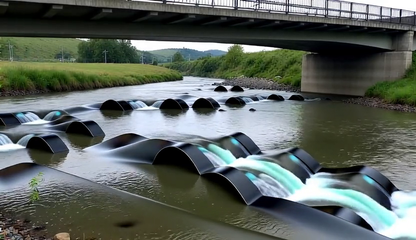}} \\
\vspace{-5mm} \\

\bmvaHangBox{\includegraphics[width=0.2\linewidth]{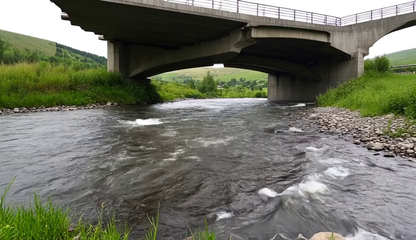}} &
\bmvaHangBox{\includegraphics[width=0.2\linewidth]{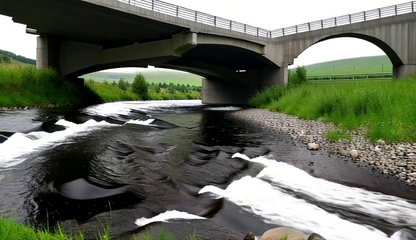}} &
\bmvaHangBox{\includegraphics[width=0.2\linewidth]{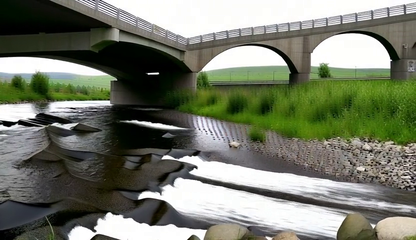}}

\end{tabular}
\caption{\textbf{Primary challenges in long-horizon video generation.}
Visual comparison of two failure modes exhibited by current AR models under fixed-camera prompts: \textbf{(1)~Motion Stagnation} (top row), where fluid motion such as a river current progressively loses temporal momentum and converges toward a static texture by $t{=}60$\,s; and \textbf{(2)~Background Drift} (bottom row), where stationary scene elements
and spatial structures gradually warp or shift over extended rollouts.}
\label{fig:generation_problems}
\end{figure*}

The fixed-camera setting is a particularly well-suited testbed for studying this trade-off. When the viewpoint is stationary, spatial drift and deliberate scene change are cleanly separable: background regions must remain geometrically stable, while dynamic regions such as water, floodwater, waves, rain, snow, fire, foliage, smoke, wind, and storms~\cite{holynski2021animating, mahapatra2022controllable} must sustain continuous motion. 
This yields two more clearly separable evaluation objectives than in moving-camera settings, making fixed-camera generation a controlled environment for studying the stability--motion trade-off. It also directly enables practical applications such as real-time ambient media, procedural game environments, and dynamic background synthesis, where extended, coherent, motion-persistent streams from a fixed viewpoint are required.

Prior works address either spatial drift~\cite{infinite-forcing, liu2025rolling, yang2025longlive, chen2026grounded} or motion decay~\cite{lu2025reward, yin2025causvid, huang2025selfforcing, zhu2026causal} in isolation; no method targets their simultaneous interaction in fixed-camera long-horizon natural flow generation. We introduce \textbf{Steady-Forcing}, a unified training and inference framework that mitigates this stability--motion trade-off through a task-specific dual-memory policy, specializing autoregressive video diffusion for static-scene nature streams. Our main
contributions are:
\begin{enumerate}

\item We characterize and empirically isolate the \textit{stability--motion trade-off} in long-horizon fixed-camera video diffusion, where stronger spatial anchoring can reduce drift but may suppress dynamic flow, while motion-rich rollouts can
inflate apparent motion through background instability.

\item We propose a \textbf{Unified Dual-Sink Mechanism} that separates persistent scene identity (V-Sink) from dynamic kinetic memory (EMA-Sink), enabling a bounded-memory context that separately preserves scene identity and compressed kinetic history during multi-minute rollouts.

\item We introduce a \textbf{Periodic KV Flush} strategy that resets the autoregressive cache at regular intervals, suppressing accumulated prediction errors before they stabilize into repeated texture artifacts.

\item We present a ground-truth-video-free, task-specialized distillation pipeline that combines a 21{,}000-prompt synthetic corpus with motion-rewarded prior initialization and a Wan2.1-14B teacher to specialize a general-purpose AR model for fixed-camera nature streams without ground-truth video supervision.

\end{enumerate}
%

Experiments demonstrate that Steady-Forcing substantially reduces background drift while maintaining visually plausible flow dynamics over multi-minute rollouts. Across seven forcing-based baselines~\cite{yin2025causvid, huang2025selfforcing, infinite-forcing, liu2025rolling, lu2025reward, zhu2026causal, yang2025longlive}, it improves long-horizon background consistency and visual quality, while a blind user study shows higher perceived motion continuity.


\section{Related Work}
\label{sec:related}

\textbf{Video Diffusion and DiTs.}
Early video diffusion models~\cite{diffusion23, ho2022video, ho2022imagen, singer2022make, blattmann2023align} denoised all frames simultaneously using Space-Time U-Nets. The field has converged on Diffusion Transformer~\cite{peebles2023scalabledit} backbones for better scaling. CogVideoX~\cite{yang2025cogvideox} and Wan2.1/2.2~\cite{wan2025} established strong quality baselines on DiT architectures, with Wan2.2 introducing a Mixture-of-Experts formulation that expands capacity without increasing inference cost. These bidirectional models rely on full future temporal context, precluding low-latency streaming inference.

\textbf{Autoregressive Long Video Generation.}
To enable streaming inference, recent work distills bidirectional teachers into few-step causal students~\cite{kodaira2025streamdit}. CausVid~\cite{yin2025causvid} established the asymmetric DMD distillation pipeline. Self-Forcing~\cite{huang2025selfforcing} bridges the train-test gap via full AR self-rollout~\cite{chen2024diffusion} during training, achieving 17 FPS real-time generation. Rolling Forcing~\cite{liu2025rolling} reduces error accumulation by jointly denoising overlapping frames ~\cite{ruhe2024rolling, kim2024fifo, wang2023gen, qiu2024freenoise, lu2024freelong}. Causal Forcing~\cite{zhu2026causal} uses an AR teacher for ODE initialization, surpassing Self-Forcing by 19.3\% in Dynamic Degree. LongLive~\cite{yang2025longlive} extends the AR design to interactive multi-minute generation via a KV-recache
mechanism~\cite{gao2024vid, wang2024loong, teng2025magi}. However, these methods either trade motion dynamics for spatial stability or accumulate background drift, a trade-off that is particularly pronounced in fixed-camera nature scenes.

\textbf{Unbounded Horizon Extension.}
Infinity-RoPE~\cite{yesiltepe2025infinity} addresses the hard temporal limit of 3D-RoPE~\cite{su2024ropeformer} without retraining, reformulating temporal encoding as a moving local reference frame. Newly generated latent blocks are indexed relative to the model's maximum horizon while earlier blocks are rotated backward, with KV Flush and RoPE Cut operators enabling prompt responsiveness and scene transitions. While this and related approaches~\cite{li2025stable, helios} extend generation length, they do not address the stability--motion trade-off for fixed-camera natural flows.

\textbf{Motion-Enhanced Distillation.}
Reward Forcing~\cite{lu2025reward} tackles motion stagnation in distilled streaming models via two contributions we directly build on: EMA-Sink, which fuses evicted frames into a dynamically updated global context via exponential moving average; and Re-DMD, which reweights the distillation objective toward high-reward dynamic regions rated by a vision-language model~\cite{liu2026improving_videoalign}, achieving 23.1 FPS with an 88.38\% improvement in dynamic amplitude. Steady-Forcing specializes their interaction for the stability--motion trade-off in fixed-camera nature streams.

\textbf{Video Generation Evaluation.}
VBench~\cite{huang2023vbench} introduced 16 disentangled evaluation dimensions, including Background Consistency and Dynamic Degree, validated against human preference annotations, which we adopt as our primary metrics. However, neither dimension exposes long-horizon flow decay in static-scene generation: Dynamic Degree measures aggregate motion amplitude rather than long-horizon persistence or late-stage flow decay, and Background Consistency does not penalize stability achieved by suppressing motion. These blind spots motivate our evaluation protocol in Section~\ref{sec:quant_res}.


\section{Background}
\label{sec:bg}

\textbf{Base Model: Wan2.1}
Steady-Forcing builds on the Wan2.1-T2V~\cite{wan2025} architecture, a flow-matching DiT operating in a compressed latent space. A causal 3D Variational Autoencoder encodes video at $4\times$ temporal and $8\times$ spatial compression, substantially reducing token count for transformer attention. The diffusion process follows Rectified Flow, in which a neural velocity field $\mathbf{v}_\theta$ interpolates between Gaussian noise and clean latents along straight ODE trajectories. We use the 1.3B parameter variant~\cite{wan2025} as the student backbone for training feasibility, and
the 14B parameter variant as the frozen teacher to provide a stronger motion prior than the 1.3B teachers used by prior forcing-based methods~\cite{huang2025selfforcing, lu2025reward}.

\textbf{3D Rotary Positional Embedding (3D-RoPE).}
Wan2.1~\cite{wan2025} employs 3D-RoPE~\cite{su2024ropeformer} to encode the temporal ($f$), height ($h$), and width ($w$) coordinates of each token via rotation matrices applied to query and key projections. Each positional dimension is trained up to 1024 temporal indices. When autoregressive rollouts extend beyond this horizon, the resulting positional representations become out-of-distribution: the RoPE formulation remains
mathematically valid, but attention weights were never optimized for such indices, leading to progressive temporal attention degradation. Steady-Forcing addresses this via Block-Relativistic RoPE~\cite{yesiltepe2025infinity}, which reformulates temporal encoding as a moving local reference frame so that indices remain within the trained range regardless of generation length (Section~\ref{sec:method_rope}).

\textbf{Self-Forcing DMD Distillation.}
The training loop of Steady-Forcing follows the Self-Forcing~\cite{huang2025selfforcing} paradigm, which bridges the train-test distribution gap through AR self-rollout during training. Unlike standard training on ground-truth context frames, which never exposes the model to its own prediction errors, Self-Forcing conditions each frame on the model's own previously generated outputs, forcing recovery from the compounding errors encountered at inference and thereby mitigating exposure bias. Steady-Forcing inherits this training loop rather than extending it to longer unrolls (as in Self-Forcing++~\cite{cui2025self_pp}), achieving long-horizon stability through the Dual-Sink memory architecture at lower training cost. A video-level Distribution Matching Distillation (DMD)~\cite{yin2024onestep_dmd, yin2024improved_dmd}, following the broader few-step diffusion distillation paradigm~\cite{salimans2022progressive, song2023consistency, luo2023latent, meng2023distillation, sauer2024adversarial}, aligns the student distribution $p_{\theta}$ with the teacher distribution $p_{\mathrm{data}}$ across the full rollout, supervised at the sequence level rather than frame-by-frame.


\section{Methodology}
\label{sec:method}

Steady-Forcing targets the stability--motion trade-off in long-horizon, fixed-camera nature video generation. We formalize the two failure modes as follows. \textbf{Drift} is the progressive geometric displacement of static background regions over autoregressive rollout time, which can be estimated by feature-aligned background displacement between the initial and current frame. \textbf{Stagnation} is the progressive decay in mean optical-flow magnitude within dynamic foreground regions. The trade-off arises when a method reduces drift by suppressing the model's sensitivity to temporal change, at the cost of accelerating stagnation, and vice versa. Our goal is to decouple these two objectives through separate memory pathways, rather than forcing them to share a single attention context.

\subsection{Unified Dual-Sink Mechanism}
\label{sec:method_dualsink}
The core memory design of Steady-Forcing separates spatial and kinetic information into two distinct attention sink components, combined within a single global context. The \textbf{V-Sink} addresses drift by providing an immutable spatial reference. The \textbf{EMA-Sink} addresses stagnation by maintaining a compressed, continuously updated summary of recent motion. Together, they form a constant-memory attention context that the model attends to at every generation step.

\begin{figure*}[t]
\centering
\includegraphics[scale=0.2]{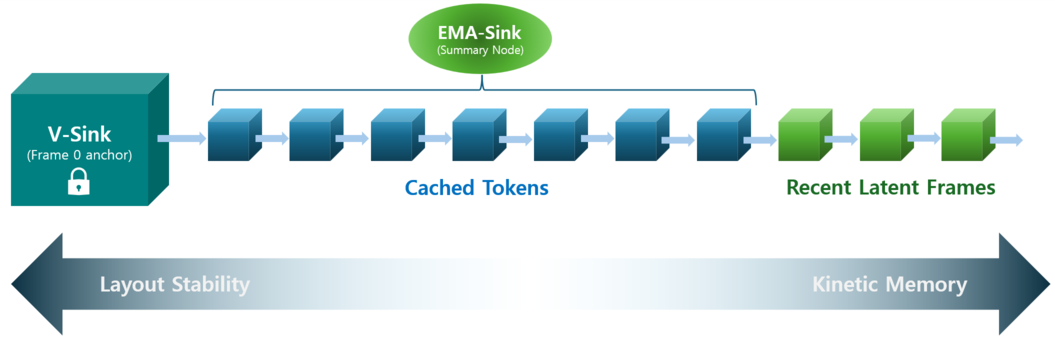}
\caption{Unified Dual-Sink Mechanism. Steady-Forcing decouples spatial persistence from kinetic memory using two complementary attention sinks. The \textbf{V-Sink} (Frame~0 keys and values) is permanently retained in the KV cache, providing a fixed spatial reference that anchors background identity throughout the rollout. The \textbf{EMA-Sink} maintains a compressed summary of motion dynamics: as frames exit the sliding window, their key-value pairs are fused into the global context via exponential moving average rather than discarded, preserving fluid momentum without growing memory cost. Attending to both sinks simultaneously allows the model to maintain background layout stability and motion persistence over extended rollouts.}
\label{fig:dual_sink}
\end{figure*}

\textbf{V-Sink (Spatial Anchor).}
\label{sec:method_vsink}
Following Infinite-Forcing~\cite{infinite-forcing}, we retain the key-value pairs of the entire initial frame (Frame~0) as a permanent global anchor $S_{\text{fixed}}$ in the KV cache throughout the rollout. Unlike token-level attention sinks in language models~\cite{xiao2024efficient}, which act as soft bias absorbers, the V-Sink serves as a \emph{spatial} reference: it holds the precise layout, color distribution, and structural identity of the scene at generation onset. Because $S_{\text{fixed}}$ is never evicted or updated, the model can always attend to the original scene state, directly counteracting the background displacement that accumulates from exposure bias during long self-rollouts. These tokens carry no temporal position conflict under Block-Relativistic RoPE because Frame~0 is always assigned index~0. The V-Sink alone, however, provides no mechanism for preserving motion: a model attending primarily to a static anchor can still suffer from motion stagnation, where dynamic textures solidify and large-scale flows collapse into weak local motion. This motivates the EMA-Sink.

\textbf{EMA-Sink (Kinetic Memory).}
\label{sec:method_emasink}
To complement the static V-Sink and prevent motion stagnation, we integrate the EMA-Sink proposed by Reward-Forcing~\cite{lu2025reward}. Rather than discarding key-value pairs when frames exit the sliding attention window of size $w$, we fuse them into a compressed global state $S_i$ via exponential moving average (EMA):
\begin{equation}
\label{eq:ema}
\begin{aligned}
S_i^K &= \alpha \cdot S_{i-1}^K + (1-\alpha) \cdot K_{i-w} \\
S_i^V &= \alpha \cdot S_{i-1}^V + (1-\alpha) \cdot V_{i-w}
\end{aligned}
\end{equation}
where $K_{i-w}$ and $V_{i-w}$ are the key and value tensors of the frame being evicted, $S_{i-1}^K$ and $S_{i-1}^V$ are the previous compressed states, and $\alpha \in (0,1)$ is the momentum decay factor (set to $\alpha = 0.99$ for both training and inference, following~\cite{lu2025reward}). 
The resulting $S_i$ acts as a coarse-grained, exponentially weighted summary of the motion history outside the local window: older evicted frames are gradually down-weighted, allowing the global memory to retain long-range kinetic context without growing in size.
The EMA-Sink operates independently of the V-Sink: $S_{\text{fixed}}$ is never updated, while $S_i$ is refreshed at every generation step.

\textbf{Block-Relativistic RoPE (Extended Horizon).}
\label{sec:method_rope}
As described in Section~\ref{sec:bg}, the 3D-RoPE in Wan2.1~\cite{wan2025} is trained for a maximum of 1024 temporal indices. Generating beyond the training horizon causes temporal attention degradation because positional representations become increasingly out-of-distribution. We integrate Block-Relativistic RoPE from Infinity-RoPE~\cite{yesiltepe2025infinity}, which resolves this by treating temporal encoding as a \emph{moving local reference frame}: each newly generated latent block is assigned indices relative to the model's maximum horizon, while earlier cached blocks are rotated backward to preserve relative temporal geometry. This eliminates fixed absolute positions and allows autoregressive generation to proceed beyond the trained absolute index range.

Our integration extends the original formulation in one respect: the V-Sink and EMA-Sink tokens receive fixed positional assignments independent of the rolling frame counter. The V-Sink (Frame~0) retains temporal index 0 throughout the rollout, preserving its role as a stable spatial reference regardless of how many frames have been generated. The EMA-Sink is assigned a fixed intermediate index that separates it from both the spatial anchor and the local window, preventing positional overlap between the three context
components.

\textbf{Unified Global Context.}
At each generation step $i$, the model attends to the concatenated global context:
\begin{equation}
\label{eq:context}
\begin{aligned}
K_i^{\text{global}} &= \bigl[S_{\text{fixed}}^K \;;\; S_i^K \;;\; K_{i-w+1:i}\bigr] \\
V_i^{\text{global}} &= \bigl[S_{\text{fixed}}^V \;;\; S_i^V \;;\; V_{i-w+1:i}\bigr]
\end{aligned}
\end{equation}
where $S_{\text{fixed}}$ is the permanent V-Sink (Frame~0), $S_i$ is the dynamically updated EMA-Sink, and $K/V_{i-w+1:i}$ are the key/value pairs of the $w$ most recent latent frames in the local sliding window. This three-part context assigns each memory role to a distinct component: long-range scene identity to $S_{\text{fixed}}$, medium-range motion history to $S_i$, and short-range fine-grained temporal continuity to the local window. The total memory footprint is $\mathcal{O}(w + s)$, where $w$ is the window size and $s$ is the combined sink size, remaining constant as generation length grows.

\subsection{Periodic KV Flush (Cache Purification)}
\label{sec:method_flush}
Long AR rollouts accumulate prediction errors in the KV cache: minor per-frame deviations in static regions compound across steps, eventually producing repeated texture artifacts, locally over-sharpened patterns that resist further change and spread across previously dynamic areas. We adapt the KV Flush operator from Infinity-RoPE~\cite{yesiltepe2025infinity} into a \emph{periodic} cache purification strategy. Rather than flushing only in response to a prompt change (as in~\cite{yesiltepe2025infinity}), we trigger the reset at regular intervals so that cache contamination is cleared before it can stabilize into persistent visual artifacts.

\begin{figure*}[t]
\centering
\includegraphics[scale=0.2]{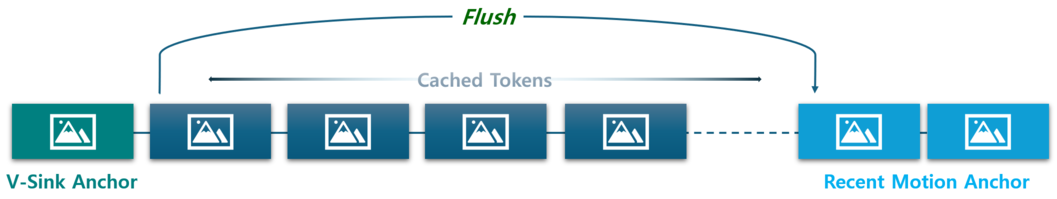}
\caption{Periodic KV Flush. Every $N_{\text{purify}}=21$ blocks, the KV cache is reset to a minimal context: the V-Sink (Frame~0 keys and values) for spatial continuity, and the $m=5$ most recent latent frames for local temporal coherence. This discards accumulated cache errors that would otherwise manifest as repeated texture artifacts in static regions over long rollouts, while the retained anchors prevent perceptual discontinuity at the reset boundary (Section~\ref{sec:method_flush}).}
\label{fig:kv_flush}
\end{figure*}

\textbf{Flush trigger.} Let $i$ denote the current block index in the AR generation sequence, where each block comprises one latent chunk of $\Delta$ frames. The flush is triggered periodically:
\begin{equation}
\label{eq:flush_trigger}
    i \bmod N_{\text{purify}} = 0, \quad N_{\text{purify}} = 21
\end{equation}

We set $N_{\text{purify}} = 21$ blocks for an optimal balance. Intervals shorter than that disrupt motion continuity by flushing before meaningful dynamics accumulate, while longer intervals allow cache errors to stabilize.

\textbf{Cache reset.} At the trigger index, the full KV cache $C_i$ is pruned to a minimal context:
\begin{equation}
\label{eq:flush_reset}
    C_i \leftarrow \bigl[KV_{\text{anchor}} \;;\; KV_{i-m+1:i}\bigr]
\end{equation}
%

where $KV_{\text{anchor}} = S_{\text{fixed}}$ is the permanent V-Sink (Frame~0 keys and values), $KV_{i-m+1:i}$ are the key-value pairs of the $m=5$ most recent latent frames, and all intermediate blocks are discarded. Retaining the V-Sink preserves scene identity across the reset boundary, while retaining the most recent blocks preserves local temporal continuity. We also discard the accumulated EMA-Sink at each flush: although it preserves motion history during normal rollout, after many autoregressive steps it may contain drifted or artifact-contaminated dynamics. Resetting it allows the EMA-Sink to rebuild from clean recent context $m$, consistent with the Block-Relativistic RoPE re-anchoring at the reset boundary.


\subsection{Rewarded DMD Distillation with Self-Forcing Unroll}
\label{sec:method_training}
Steady-Forcing specializes an autoregressive student for fixed-camera natural flow generation through two training components: motion-biased weight initialization and Self-Forcing DMD distillation~\cite{huang2025selfforcing} with domain-specific negative prompting.

\textbf{Motion-Prior Initialization.}
\label{sec:method_init}
We initialize the student generator with weights from Reward-Forcing~\cite{lu2025reward} rather than from the original Wan2.1~\cite{wan2025} checkpoint. Reward-Forcing's Re-DMD training reweights the distillation objective toward high-reward dynamic regions, producing a checkpoint with substantially higher motion amplitude than standard DMD initialization. This matters for our setting because DMD distillation of nature-scene content from a neutral initialization tends to converge toward a low-energy sub-distribution: the model learns to produce spatially stable outputs that score well on frame-level quality metrics but suppress fluid motion. Starting from the Reward-Forcing checkpoint biases the student's initial
distribution toward high-amplitude dynamics, providing a stronger prior for the fluid physics the model must sustain. 
We qualitatively examine the impact of this initialization choice in the stage-wise ablation study (Section~\ref{sec:ablation}).

\textbf{Self-Forcing DMD with Domain-Specific Unroll.}
\label{sec:method_unroll}
We distill from a frozen Wan2.1-T2V-14B~\cite{wan2025} teacher under
the Self-Forcing training loop~\cite{huang2025selfforcing}. During
training, the student generates video chunk-by-chunk conditioning on
its own previous outputs, with V-Sink, EMA-Sink, Block-Relativistic
RoPE, and the local sliding window all active — matching the inference
configuration exactly. A video-level DMD~\cite{yin2024onestep_dmd,
yin2024improved_dmd} loss aligns the student rollout distribution with
the teacher distribution; the full objective derivation is provided in
the supplementary material.

\textbf{Expanded negative prompting.} To suppress domain-specific
failure modes, we append a fixed set of negative descriptors to the
CFG~\cite{ho2022classifier_cfg} negative prompt during both distillation
and inference, such as: \textit{``repetitive round textures, ground artifacts, motion stagnation, frozen motion, static water, hardened flow.''}

\section{Experiments}
\label{sec:experiments}

\subsection{Implementation Details}
\label{sec:impl}

We implement Steady-Forcing on the Wan2.1-T2V-1.3B~\cite{wan2025} backbone, a flow-matching DiT~\cite{peebles2023scalabledit}, distilling from a frozen Wan2.1-T2V-14B teacher, a stronger motion prior than the 1.3B teachers used in prior forcing-based methods~\cite{huang2025selfforcing, lu2025reward}. The student is initialized from the Reward-Forcing~\cite{lu2025reward} checkpoint, which Lu et al. report achieving an 88.38\% improvement in dynamic amplitude on their benchmark via Re-DMD; we use this checkpoint for its motion-biased weight prior rather than neutral ODE initialization.

\textbf{Training schedule.} We follow the Self-Forcing DMD protocol~\cite{huang2025selfforcing} with a uniform 4-step denoising schedule and train for 6{,}000 iterations (batch size~8). The AdamW optimizer uses a generator learning rate of $2.0\times10^{-6}$ and a critic learning rate of $4.0\times10^{-7}$. 
We use 3 frames per block, with the periodic KV-flush interval set to $N_{\text{purify}}=21$ blocks (Section~\ref{sec:method_flush}).

\textbf{Ground-truth-video-free prompt corpus.} To avoid dependence on curated video datasets, we synthesize a training corpus of 21{,}000 prompts by combinatorially sampling from semantic pools covering static scene anchors, fluid flow types, atmospheric conditions, and global locations. Each prompt pairs fixed environmental elements with high-amplitude dynamic descriptors and appends a camera constraint specifying a stationary, tripod-mounted viewpoint, encouraging the student to learn background stability and fluid dynamics jointly without ground-truth supervision.

\textbf{Evaluation prompts.}
We prepare a held-out evaluation set using an LLM-assisted prompt generation process followed by manual filtering, focusing on fixed-camera, continuous nature-flow scenarios. The evaluation prompts are disjoint from the 21{,}000-prompt synthetic training corpus, and the full prompt list is provided in the supplementary material. The set is organized into four horizon tiers: T1 (5s, $n{=}6$), T2 (60s, $n{=}23$), T3 (120s, $n{=}6$), and T4 (240s, $n{=}4$). 


Tables~\ref{tab:quant_res_short_med} and~\ref{tab:quant_res_long_extended} report quantitative results across the four horizon tiers. Representative qualitative comparison uses T2 prompts (Fig.~\ref{fig:qual_res}), and extreme-horizon failure modes are shown on T4 prompts (Fig.~\ref{fig:limitation_results}).

\textbf{Hardware.} All experiments run on 8$\times$ NVIDIA A100 (80\,GB) GPUs. Total training time is approximately 67 hours. Inference is run on a single GPU.

\subsection{Qualitative Results}
\label{sec:qual_res}

\begin{figure*}[t]
\centering
\setlength{\tabcolsep}{1pt} 
\begin{tabular}{rcccccccc}

& 
{\scriptsize $^\dagger$CausVid} & 
{\scriptsize \begin{tabular}{c}$^\dagger$Self-\\Forcing\end{tabular}} & 
{\scriptsize \begin{tabular}{c}$^\dagger$Infinite-\\Forcing\end{tabular}} & 
{\scriptsize \begin{tabular}{c}$^\dagger$Rolling-\\Forcing\end{tabular}} & 
{\scriptsize \begin{tabular}{c}$^\dagger$Reward-\\Forcing\end{tabular}} & 
{\scriptsize $^\dagger$LongLive} & 
{\scriptsize \begin{tabular}{c}$^\dagger$Causal-\\Forcing\end{tabular}} & 
{\tiny \begin{tabular}{c}\textbf{Steady-}\\\textbf{Forcing}\\\textbf{(Ours)}\end{tabular}} \\ [0.3em]

\scriptsize $t=0\text{s}$ &
\bmvaHangBox{\includegraphics[width=0.11\linewidth]{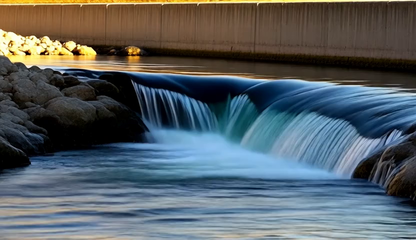}} &
\bmvaHangBox{\includegraphics[width=0.11\linewidth]{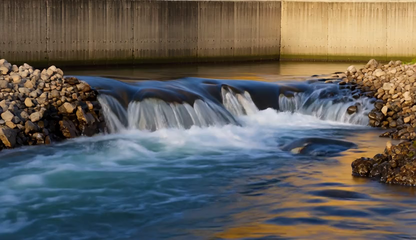}} &
\bmvaHangBox{\includegraphics[width=0.11\linewidth]{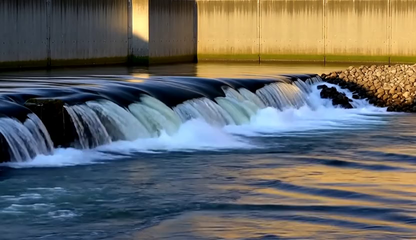}} &
\bmvaHangBox{\includegraphics[width=0.11\linewidth]{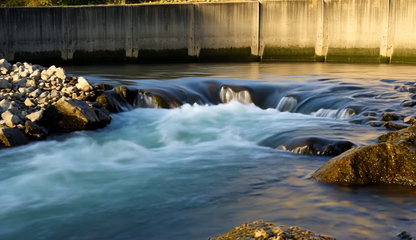}} &
\bmvaHangBox{\includegraphics[width=0.11\linewidth]{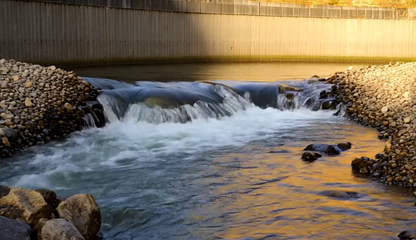}} &
\bmvaHangBox{\includegraphics[width=0.11\linewidth]{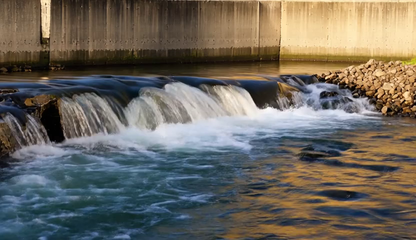}} &
\bmvaHangBox{\includegraphics[width=0.11\linewidth]{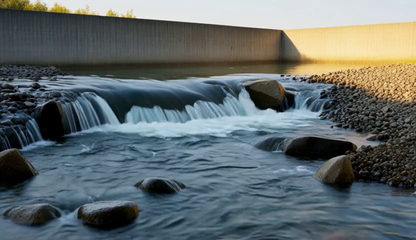}} &
\bmvaHangBox{\includegraphics[width=0.11\linewidth]{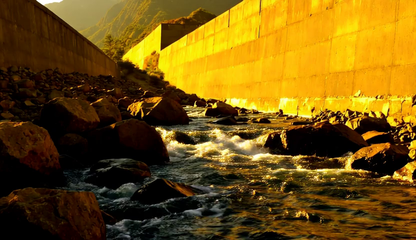}} \\
\vspace{-5mm} \\

\scriptsize $t=20\text{s}$ &
\bmvaHangBox{\includegraphics[width=0.11\linewidth]{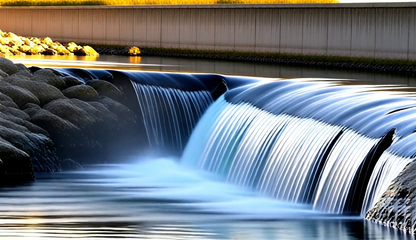}} &
\bmvaHangBox{\includegraphics[width=0.11\linewidth]{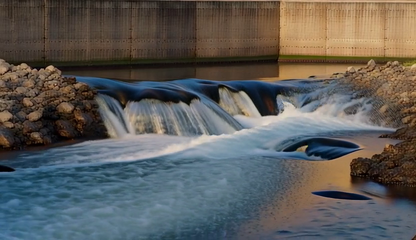}} &
\bmvaHangBox{\includegraphics[width=0.11\linewidth]{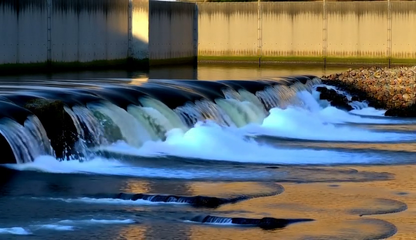}} &
\bmvaHangBox{\includegraphics[width=0.11\linewidth]{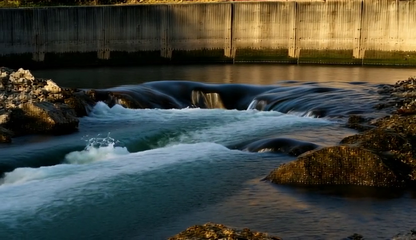}} &
\bmvaHangBox{\includegraphics[width=0.11\linewidth]{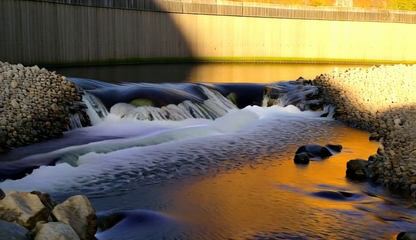}} &
\bmvaHangBox{\includegraphics[width=0.11\linewidth]{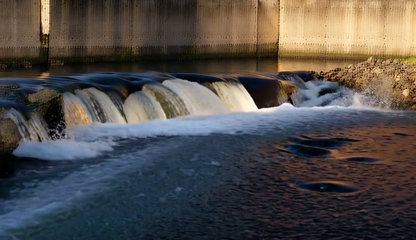}} &
\bmvaHangBox{\includegraphics[width=0.11\linewidth]{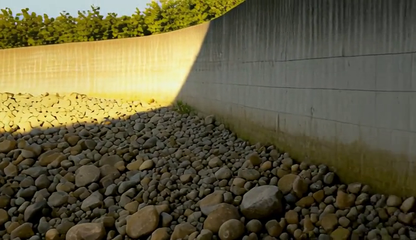}} &
\bmvaHangBox{\includegraphics[width=0.11\linewidth]{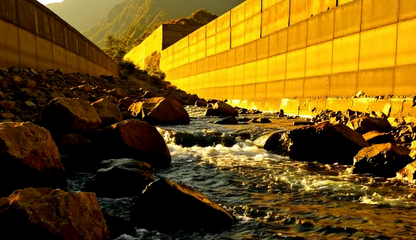}} \\
\vspace{-5mm} \\

\scriptsize $t=40\text{s}$ &
\bmvaHangBox{\includegraphics[width=0.11\linewidth]{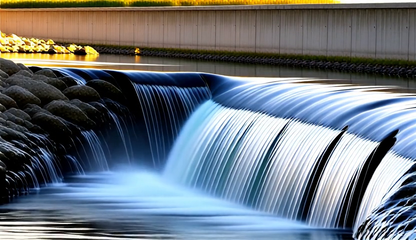}} &
\bmvaHangBox{\includegraphics[width=0.11\linewidth]{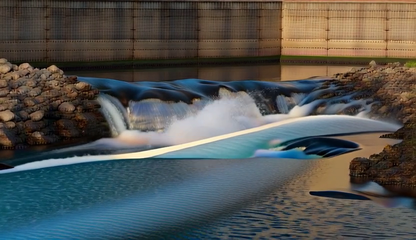}} &
\bmvaHangBox{\includegraphics[width=0.11\linewidth]{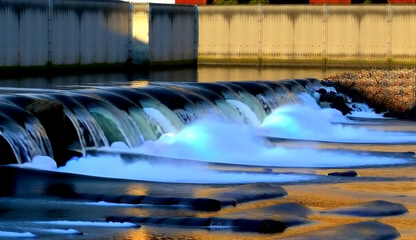}} &
\bmvaHangBox{\includegraphics[width=0.11\linewidth]{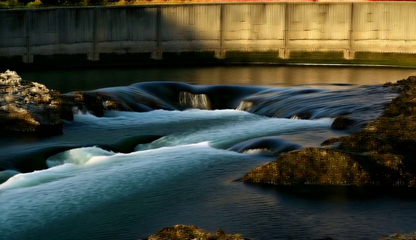}} &
\bmvaHangBox{\includegraphics[width=0.11\linewidth]{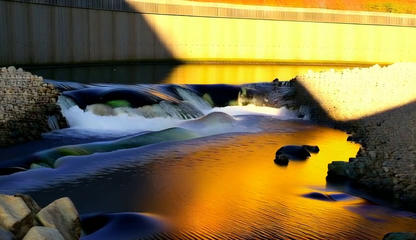}} &
\bmvaHangBox{\includegraphics[width=0.11\linewidth]{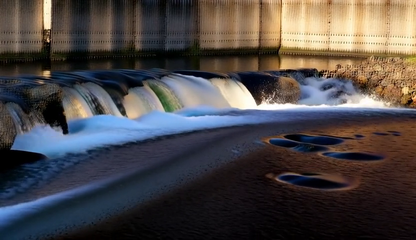}} &
\bmvaHangBox{\includegraphics[width=0.11\linewidth]{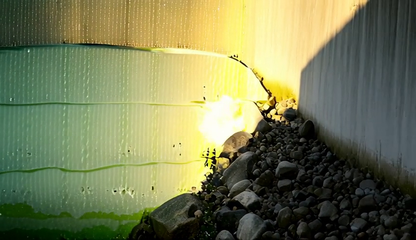}} &
\bmvaHangBox{\includegraphics[width=0.11\linewidth]{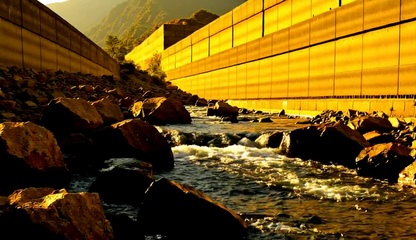}} \\
\vspace{-5mm} \\

\scriptsize $t=60\text{s}$ &
\bmvaHangBox{\includegraphics[width=0.11\linewidth]{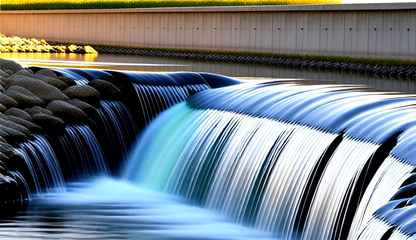}} &
\bmvaHangBox{\includegraphics[width=0.11\linewidth]{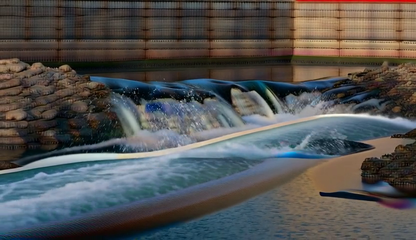}} &
\bmvaHangBox{\includegraphics[width=0.11\linewidth]{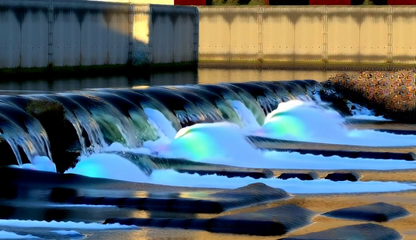}} &
\bmvaHangBox{\includegraphics[width=0.11\linewidth]{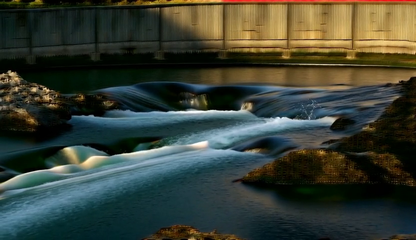}} &
\bmvaHangBox{\includegraphics[width=0.11\linewidth]{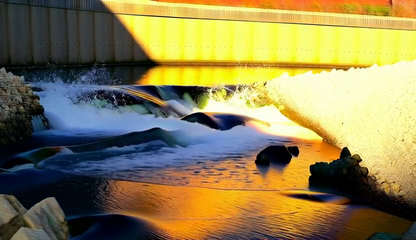}} &
\bmvaHangBox{\includegraphics[width=0.11\linewidth]{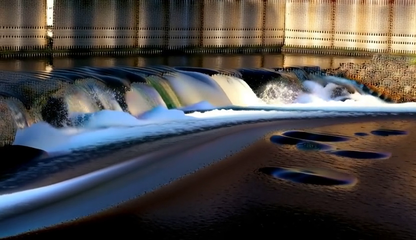}} &
\bmvaHangBox{\includegraphics[width=0.11\linewidth]{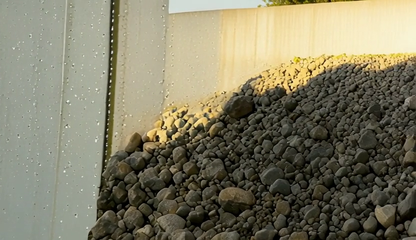}} &
\bmvaHangBox{\includegraphics[width=0.11\linewidth]{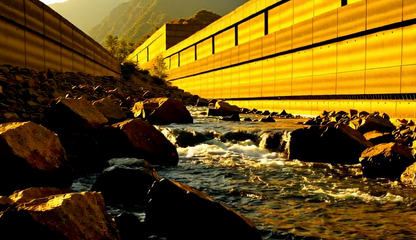}} \\

\end{tabular}
\caption{\textbf{Qualitative comparison at 60-second horizon.}
Frames are sampled at $t \in \{0, 20, 40, 60\}$\,s. Steady-Forcing better preserves spatial layout and visually plausible fluid motion over extended rollouts, while baselines exhibit increasing drift, motion decay, or texture artifacts at later timesteps. $^\dagger$~Public baseline weights evaluated under our fixed-camera inference protocol.}
\label{fig:qual_res}
\end{figure*}

We compare Steady-Forcing against seven forcing-based baselines: CausVid~\cite{yin2025causvid}, Self-Forcing~\cite{huang2025selfforcing}, Infinite-Forcing~\cite{infinite-forcing}, Rolling-Forcing~\cite{liu2025rolling},
Reward-Forcing~\cite{lu2025reward}, LongLive~\cite{yang2025longlive}, and Causal-Forcing~\cite{zhu2026causal}. For daggered baselines, we evaluate public pretrained checkpoints under the same fixed-camera steady-motion inference wrapper used for our method (V-Sink, EMA-Sink, KV Flush Loop and Block-Relativistic RoPE). These daggered results should therefore be interpreted as task-adapted variants rather than native release scores (Fig.~\ref{fig:qual_res}).

\textbf{Drift mitigation.}
Baselines such as CausVid~\cite{yin2025causvid} and Self-Forcing~\cite{huang2025selfforcing} progressively accumulate exposure bias, manifesting as background translation and color drift over time. The V-Sink (Frame~0 anchor) visibly reduces this displacement and helps maintain background layout consistency across multi-minute rollouts.

\textbf{Flow persistence.}
Reward-Forcing~\cite{lu2025reward} tends to collapse toward reduced-motion states in scenes where rewarded dynamic regions represent a small spatial fraction. The EMA-Sink (Section~\ref{sec:method_emasink}) counteracts this by continuously updating the global context with evicted frame dynamics, preserving downstream flow in directional scenes such as rivers and canals.

\textbf{Artifact suppression.}
At longer horizons, smaller AR models commonly produce over-stabilized textures, locally hardened patterns resembling carved rock or metallic surfaces. The Periodic KV Flush (Section~\ref{sec:method_flush}) prevents these artifacts from stabilizing by clearing accumulated cache errors at regular intervals, maintaining a natural, fluid-like appearance throughout the rollout.


\subsection{Quantitative Results}
\label{sec:quant_res}

We evaluate all methods on six VBench~\cite{huang2023vbench} dimensions:
\textit{Background Consistency}, \textit{Motion Smoothness}, \textit{Imaging Quality}, \textit{Temporal Flickering}, \textit{Aesthetic Quality}, and \textit{Dynamic Degree}. Subject Consistency is excluded as our prompts contain no foreground subject. 

Dynamic Degree warrants careful interpretation in a fixed-camera setting: optical flow from camera drift and from genuine fluid motion are both rewarded equally, so high scores can reflect spatial instability rather than desired flow. We therefore interpret it alongside Background Consistency and the human preference study rather than as an isolated indicator of desired motion; a method that suppresses camera drift necessarily produces less background optical flow, which VBench registers as lower Dynamic Degree regardless of fluid motion quality.




Tables~\ref{tab:quant_res_short_med} and~\ref{tab:quant_res_long_extended} show that Steady-Forcing achieves the highest Background Consistency and Imaging Quality at every reported horizon: 98.06/69.87 at 5s, 95.60/71.59 at 60s, 95.07/66.21 at 120s, and 92.57/71.27 at 240s. Background Consistency generally degrades more slowly for Steady-Forcing than for the baselines, which is consistent with the V-Sink's intended role of anchoring spatial layout over extended rollouts. Steady-Forcing also achieves the best or second-best Motion Smoothness, Temporal Flickering and Aesthetic Quality across all reported horizons, clearly depicting the high efficacy of the proposed pipeline.


Across all evaluated horizons, Steady-Forcing retains the highest Background Consistency and Imaging Quality, however its VBench averages are lower than Causal-Forcing, which obtains the highest Dynamic Degrees along with lower Background Consistencies (Tab.~\ref{tab:quant_res_short_med},~\ref{tab:quant_res_long_extended}). This illustrates a limitation of using VBench Avg. alone for fixed-camera nature streams: Dynamic Degree rewards all optical flow, including drift-induced background motion, while it does not directly measure whether flow remains semantically plausible or whether textures harden over time (Fig.~\ref{fig:qual_res}). We therefore avoid treating VBench Avg. as the sole ranking criterion and interpret Dynamic Degree together with Background Consistency and the preference study over representative baselines. Under this task-specific reading, Steady-Forcing targets the stable-motion regime: high background consistency, strong imaging \& aesthetic qualities, competitive temporal flickering and smoothness along with continuous dynamic motion, further supported by the user-study preference for static stability and motion continuity (Sec.~\ref{sec:user_study}).

Thus, Steady-Forcing is not optimized for the generic VBench average; it is optimized for the fixed-camera nature-flow objective, where spatial persistence and perceived motion continuity must be satisfied simultaneously.



\begin{table*}[t]
\begin{center}
\tiny
\setlength{\tabcolsep}{2pt} 
\begin{tabular}{l|ccccccc|ccccccc}
\hline
Method & \multicolumn{7}{c|}{Results on 5s $\uparrow$} & \multicolumn{7}{c}{Results on 60s $\uparrow$} \\ \cline{2-15} 
& 
\parbox{0.8cm}{\centering Background\\Consistency} & 
\parbox{0.8cm}{\centering Motion\\Smoothness} & 
\parbox{0.6cm}{\centering Imaging\\Quality} & 
\parbox{0.7cm}{\centering Temporal\\Flickering} & 
\parbox{0.7cm}{\centering Aesthetic\\Quality} & 
\parbox{0.6cm}{\centering Dynamic\\Degree} & 
\parbox{0.4cm}{\centering Avg.} &
\parbox{0.8cm}{\centering Background\\Consistency} & 
\parbox{0.7cm}{\centering Motion\\Smoothness} & 
\parbox{0.6cm}{\centering Imaging\\Quality} & 
\parbox{0.7cm}{\centering Temporal\\Flickering} & 
\parbox{0.7cm}{\centering Aesthetic\\Quality} & 
\parbox{0.6cm}{\centering Dynamic\\Degree} & 
\parbox{0.4cm}{\centering Avg.} \\ \hline
$^\dagger$CausVid \cite{yin2025causvid}            & 96.49 & 98.65 & 51.79 & 97.99 & 61.06 & 37.01 & 73.83 & 92.51 & 98.99 & 55.72 & 98.30 & \underline{63.51} & 40.93 & 74.99 \\
$^\dagger$Self-Forcing \cite{huang2025selfforcing} & 96.47 & 98.86 & 62.50 & 98.30 & 57.58 & \underline{48.02} & 76.96 & 90.55 & 98.73 & 62.26 & 98.03 & 58.11 & \underline{65.88} & \underline{78.93} \\
$^\dagger$Infinite-Forcing \cite{infinite-forcing} & \underline{97.46} & \underline{98.96} & 63.21 & \textbf{98.68} & 56.37 & 25.70 & 73.40 & \underline{92.66} & \textbf{99.16} & 56.14 & \textbf{98.88} & 59.98 & 26.46 & 72.21 \\
$^\dagger$Rolling-Forcing \cite{liu2025rolling}    & 96.63 & 98.55 & 64.69 & 97.65 & 60.81 & 47.06 & \underline{77.57} & 92.05 & 98.63 & 59.84 & 97.89 & 58.61 & 48.89 & 75.99 \\
$^\dagger$Reward-Forcing \cite{lu2025reward}       & 96.62 & 98.67 & 65.55 & 98.20 & 58.46 & 27.95 & 74.24 & 90.62 & 98.65 & 63.29 & 97.94 & 59.58 & 43.61 & 75.62 \\
$^\dagger$LongLive \cite{yang2025longlive}         & 95.88 & 98.72 & \underline{67.13} & 98.07 & 57.23 & 34.26 & 75.22 & 90.83 & 98.44 & 61.45 & 97.67 & 57.05 & 59.18 & 77.44 \\
$^\dagger$Causal-Forcing \cite{zhu2026causal}      & 95.85 & 97.95 & 63.46 & 96.16 & \underline{62.05} & \textbf{70.00} & \textbf{80.91} & 87.73 & 97.05 & \underline{68.30} & 94.04 & 55.91 & \textbf{91.37} & \textbf{82.40} \\
\textbf{Steady-Forcing (Ours)}           & \textbf{98.06} & \textbf{99.00} & \textbf{69.87} & \underline{98.53} & \textbf{63.50} & 25.22 & 75.70 & \textbf{95.60} & \underline{99.13} & \textbf{71.59} & \underline{98.44} & \textbf{64.85} & 37.43 & 77.84 \\ \hline
\end{tabular}
\end{center}
\caption{Quantitative evaluation on VBench~\cite{huang2023vbench} across short-horizon (5s) and medium-horizon (60s) video tiers. $^\dagger$ indicates public baseline weights evaluated under our fixed-camera inference protocol.}
\label{tab:quant_res_short_med}
\end{table*}

\begin{table*}[t]
\begin{center}
\tiny
\setlength{\tabcolsep}{2pt} 
\begin{tabular}{l|ccccccc|ccccccc}
\hline
Method & \multicolumn{7}{c|}{Results on 120s $\uparrow$} & \multicolumn{7}{c}{Results on 240s $\uparrow$} \\ \cline{2-15} 
& 
\parbox{0.8cm}{\centering Background\\Consistency} & 
\parbox{0.8cm}{\centering Motion\\Smoothness} & 
\parbox{0.6cm}{\centering Imaging\\Quality} & 
\parbox{0.7cm}{\centering Temporal\\Flickering} & 
\parbox{0.7cm}{\centering Aesthetic\\Quality} & 
\parbox{0.6cm}{\centering Dynamic\\Degree} & 
\parbox{0.4cm}{\centering Avg.} &
\parbox{0.8cm}{\centering Background\\Consistency} & 
\parbox{0.7cm}{\centering Motion\\Smoothness} & 
\parbox{0.6cm}{\centering Imaging\\Quality} & 
\parbox{0.7cm}{\centering Temporal\\Flickering} & 
\parbox{0.7cm}{\centering Aesthetic\\Quality} & 
\parbox{0.6cm}{\centering Dynamic\\Degree} & 
\parbox{0.4cm}{\centering Avg.} \\ \hline
$^\dagger$CausVid \cite{yin2025causvid}            & \underline{92.75} & \underline{98.97} & 56.95 & \underline{98.64} & \underline{62.06} & 20.47 & 71.64 & \underline{88.25} & 98.69 & 64.27 & 98.43 & \textbf{61.51} & 23.48 & 72.44 \\
$^\dagger$Self-Forcing \cite{huang2025selfforcing} & 88.55 & 98.86 & 57.66 & 98.45 & 51.48 & 29.93 & 70.82 & 83.70 & 98.54 & 55.73 & 98.01 & 49.70 & 46.89 & 72.10 \\
$^\dagger$Infinite-Forcing \cite{infinite-forcing} & 91.13 & \textbf{99.22} & 48.00 & \textbf{98.99} & 58.18 & 24.97 & 70.08 & 85.95 & \textbf{99.37} & 54.92 & \textbf{99.24} & 55.38 & 26.62 & 70.24 \\
$^\dagger$Rolling-Forcing \cite{liu2025rolling}    & 89.75 & 98.77 & 55.26 & 98.13 & 53.98 & 42.30 & 73.03 & 87.37 & 99.07 & 57.29 & 98.37 & 51.80 & \underline{55.88} & 74.96 \\
$^\dagger$Reward-Forcing \cite{lu2025reward}       & 87.65 & 98.63 & 52.63 & 97.70 & 54.44 & 44.96 & 72.67 & 86.17 & 98.51 & 61.25 & 97.85 & 53.16 & 52.93 & \underline{74.98} \\
$^\dagger$LongLive \cite{yang2025longlive}         & 88.50 & 98.48 & 53.32 & 97.74 & 52.80 & \underline{52.25} & 73.85 & 84.58 & 98.69 & 55.95 & 97.85 & 46.83 & 35.80 & 69.95 \\
$^\dagger$Causal-Forcing \cite{zhu2026causal}      & 88.09 & 96.68 & \underline{59.91} & 93.43 & 56.33 & \textbf{95.95} & \textbf{81.73} & 86.87 & 96.47 & \underline{66.36} & 92.37 & 54.79 & \textbf{94.75} & \textbf{81.94} \\
\textbf{Steady-Forcing (Ours)}           & \textbf{95.07} & \textbf{99.22} & \textbf{66.21} & \underline{98.64} & \textbf{62.16} & 37.99 & \underline{76.55} & \textbf{92.57} & \underline{99.34} & \textbf{71.27} & \underline{99.14} & \underline{60.76} & 21.86 & 74.16 \\ \hline
\end{tabular}
\end{center}
\caption{Quantitative evaluation on VBench~\cite{huang2023vbench} across long-horizon tiers (120s and extended 240s). $^\dagger$ indicates public baseline weights evaluated under our fixed-camera inference protocol.}
\label{tab:quant_res_long_extended}
\end{table*}


\section{Discussion}

\subsection{User Study}
\label{sec:user_study}

We conducted a blind preference study against three representative baselines: Self-Forcing~\cite{huang2025selfforcing}, Reward-Forcing~\cite{lu2025reward}, and Rolling-Forcing~\cite{liu2025rolling}, covering self-rollout training, motion-rewarded initialization, and rolling-window denoising, being the closest contenders except Causal-Forcing~\cite{zhu2026causal} (Sec.\ref{sec:quant_res}). To limit participant burden, we selected representative baselines rather than all seven quantitative baselines. For each of six prompts, participants were shown four anonymized videos (A/B/C/D, randomized per trial) and selected the best according to five criteria: overall quality, static-view stability, motion continuity, temporal consistency, and artifact-free quality. Participants also assigned a 1--5 Likert score to each video. 23 participants completed all six prompts, yielding 138 four-way comparisons per criterion.

Table~\ref{tab:user_study} shows that Steady-Forcing obtains the highest preference rate across all five criteria, with the strongest margins on Static-View Stability (0.746 vs.\ next-best 0.101) and Motion Continuity (0.710 vs.\ 0.116). 
The simultaneous lead on both criteria, which are typically in tension, is consistent with the intended stability--motion trade-off: participants preferred Steady-Forcing both for fixed-view stability and for perceived fluid motion. The Artifact-Free lead (0.710 vs.\ 0.109) aligns with the qualitative observation that cache purification suppresses the texture degradation visible in baselines at longer horizons. Steady-Forcing achieves a mean Likert rating of 4.138, compared to 2.283--2.572 for baselines.

\begin{table*}[t]
\begin{center}
\scriptsize
\setlength{\tabcolsep}{5pt} 
\begin{tabular}{lcccccc}
\hline
Model & 
\parbox{1.3cm}{\centering Overall\\Quality $\uparrow$} & 
\parbox{1.2cm}{\centering Static\\View $\uparrow$} & 
\parbox{1.3cm}{\centering Motion\\Continuity $\uparrow$} & 
\parbox{1.4cm}{\centering Temporal\\Consistency $\uparrow$} & 
\parbox{1.4cm}{\centering Artifact-\\Free $\uparrow$} & 
\parbox{1.2cm}{\centering Avg.\\Rating $\uparrow$} \\ \hline
$^\dagger$Self-Forcing \cite{huang2025selfforcing}   & 0.080 & 0.080 & 0.080 & \underline{0.109} & \underline{0.109} & 2.283 \\
$^\dagger$Reward-Forcing \cite{lu2025reward}       & \underline{0.123} & \underline{0.101} & \underline{0.116} & 0.080 & 0.080 & 2.283 \\
$^\dagger$Rolling-Forcing \cite{liu2025rolling}     & 0.072 & 0.072 & 0.094 & 0.058 & 0.101 & \underline{2.572} \\ 
\textbf{Steady-Forcing (Ours)}           & \textbf{0.725} & \textbf{0.746} & \textbf{0.710} & \textbf{0.754} & \textbf{0.710} & \textbf{4.138} \\ \hline
\end{tabular}
\end{center}
\caption{\textbf{Blind user study results.}
Preference rate: fraction of trials in which each method was selected as best for the given criterion, after resolving the randomized A/B/C/D presentation order (23 participants $\times$ 6 prompts $=$ 138 comparisons per criterion). Avg.\ Rating: mean 1--5 Likert score mapped back to each method. $^\dagger$~Public baseline weights under our fixed-camera inference protocol.}
\label{tab:user_study}
\end{table*}

\subsection{Ablation Study}
\label{sec:ablation}

\begin{figure*}[t]
\centering
\setlength{\tabcolsep}{2pt} 
\begin{tabular}{rccccc}

& 
\parbox{1.5cm}{\centering\scriptsize $^*$V-Sink} & 
\parbox{1.5cm}{\centering\scriptsize $^\dagger$EMA-Sink} & 
\parbox{1.5cm}{\centering\scriptsize $^\dagger$Dual-Sink} & 
\parbox{1.8cm}{\centering\tiny $^*$Dual-Sink + Periodic KV Flush} & 
\parbox{2.0cm}{\centering\scriptsize \textbf{Steady-Forcing (Ours)}} \\ [0.2em]

\scriptsize $t=0\text{s}$ &
\bmvaHangBox{\includegraphics[width=0.17\linewidth]{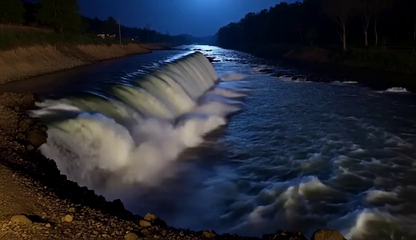}} &
\bmvaHangBox{\includegraphics[width=0.17\linewidth]{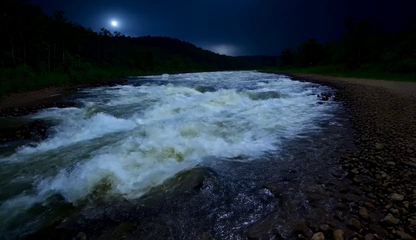}} &
\bmvaHangBox{\includegraphics[width=0.17\linewidth]{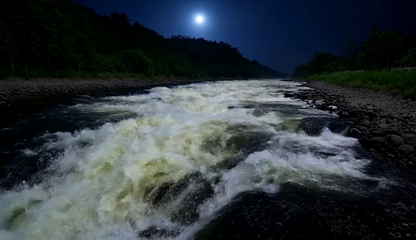}} &
\bmvaHangBox{\includegraphics[width=0.17\linewidth]{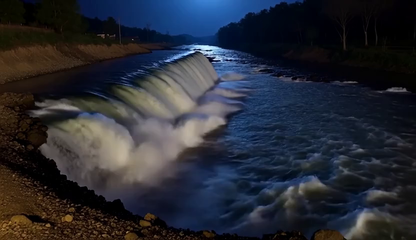}} &
\bmvaHangBox{\includegraphics[width=0.17\linewidth]{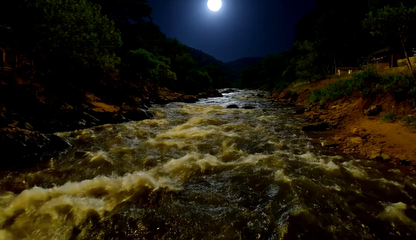}} \\
\vspace{-5mm} \\

\scriptsize $t=20\text{s}$ &
\bmvaHangBox{\includegraphics[width=0.17\linewidth]{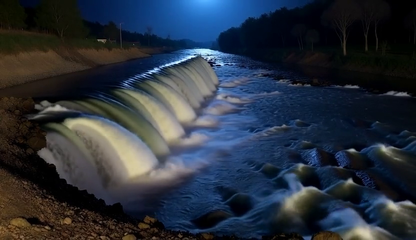}} &
\bmvaHangBox{\includegraphics[width=0.17\linewidth]{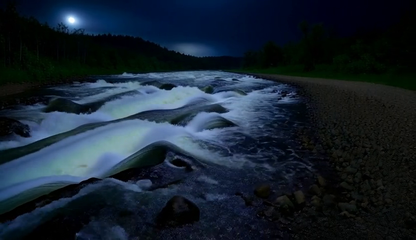}} &
\bmvaHangBox{\includegraphics[width=0.17\linewidth]{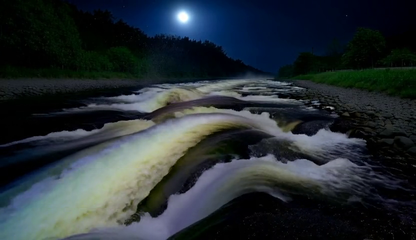}} &
\bmvaHangBox{\includegraphics[width=0.17\linewidth]{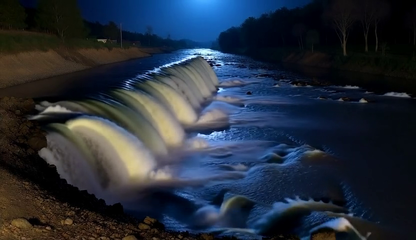}} &
\bmvaHangBox{\includegraphics[width=0.17\linewidth]{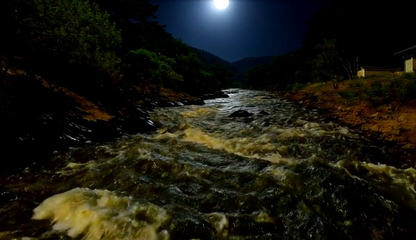}} \\
\vspace{-5mm} \\

\scriptsize $t=40\text{s}$ &
\bmvaHangBox{\includegraphics[width=0.17\linewidth]{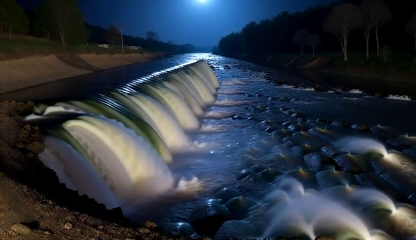}} &
\bmvaHangBox{\includegraphics[width=0.17\linewidth]{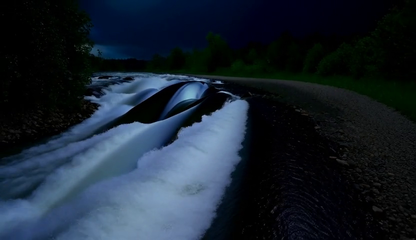}} &
\bmvaHangBox{\includegraphics[width=0.17\linewidth]{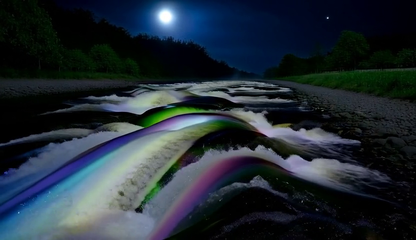}} &
\bmvaHangBox{\includegraphics[width=0.17\linewidth]{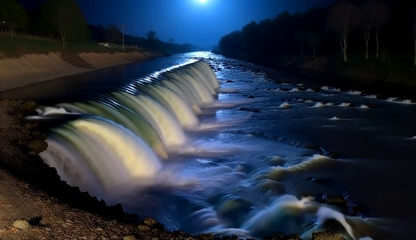}} &
\bmvaHangBox{\includegraphics[width=0.17\linewidth]{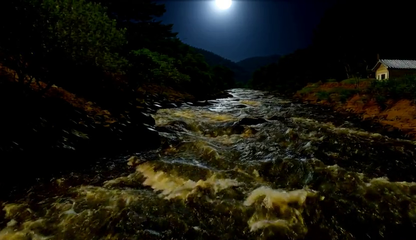}} \\
\vspace{-5mm} \\

\scriptsize $t=60\text{s}$ &
\bmvaHangBox{\includegraphics[width=0.17\linewidth]{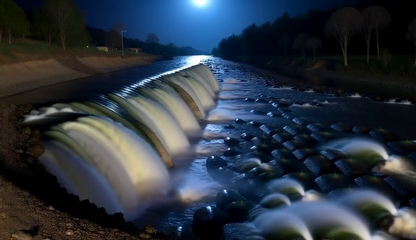}} &
\bmvaHangBox{\includegraphics[width=0.17\linewidth]{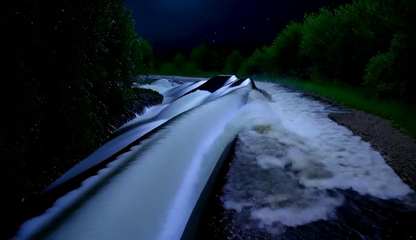}} &
\bmvaHangBox{\includegraphics[width=0.17\linewidth]{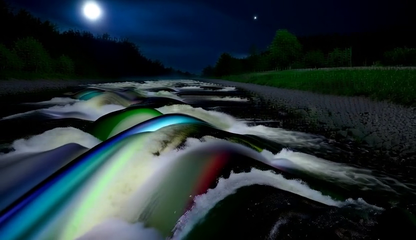}} &
\bmvaHangBox{\includegraphics[width=0.17\linewidth]{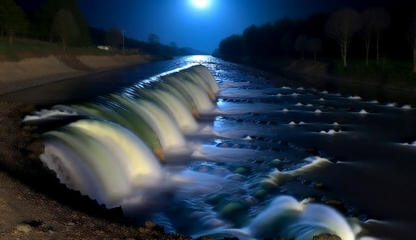}} &
\bmvaHangBox{\includegraphics[width=0.17\linewidth]{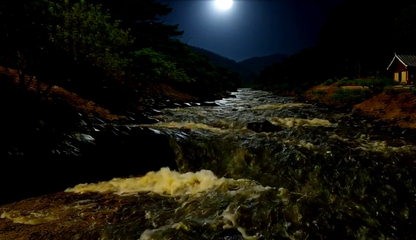}} \\

\end{tabular}
\caption{\textbf{Stage-wise ablation at 60-second horizon.}
V-Sink anchors spatial layout but permits motion stagnation; EMA-Sink preserves motion but permits background drift; adding KV Flush reduces accumulated cache artifacts. The full Steady-Forcing pipeline shows the most favorable qualitative balance between spatial layout and fluid motion throughout the rollout. $^*$~Inference on Infinite-Forcing~\cite{infinite-forcing} weights; $^\dagger$~same on Reward-Forcing~\cite{lu2025reward} weights.}
\label{fig:ablation_results}
\end{figure*}

Fig.~\ref{fig:ablation_results} summarizes stage-wise qualitative ablations at the 60-second horizon. The V-Sink alone preserves coarse spatial layout but provides no mechanism for motion preservation, leaving dynamic regions susceptible to stagnation. The EMA-Sink alone retains fluid motion history but lacks a persistent geometric anchor, allowing background drift to accumulate. Combining both sinks improves the stability--motion balance, and adding Periodic KV Flush further reduces accumulated cache artifacts during long rollouts. The Reward-Forcing initialization supplies a motion-biased starting point for the final Steady-Forcing model, while this figure focuses on the memory and cache components. Together, these qualitative results suggest that the full pipeline yields the most favorable stability--motion balance among the tested variants.

\textbf{Stability-Motion Trade-off in Practice}
The trade-off is mitigated but not fully eliminated: residual motion stagnation remains visible at longer durations, particularly in large-body water scenes where fluid dynamics exceed the EMA-Sink's compression capacity,
and mild color drift persists at the 240s horizon (Fig.~\ref{fig:limitation_results}).
These residual effects indicate remaining limits of the fixed-size memory design and the base model's training distribution.

\textbf{Efficiency Analysis}
Steady-Forcing runs at approximately 17 FPS on a single A100 GPU, comparable to Self-Forcing~\cite{huang2025selfforcing} and Rolling Forcing~\cite{liu2025rolling}. Because the V-Sink, EMA-Sink, and local window are fixed-size, KV memory remains
$\mathcal{O}(w+s)$ with respect to rollout length, although visual quality is still limited by the failure modes above.

\section{Limitations and Future Work}
\label{sec:limit}


\textbf{Inherited model constraints.}
Steady-Forcing inherits the physics priors of the Wan2.1 backbone; fluid interactions underrepresented in its training distribution can produce implausible dynamics during long rollouts.

\textbf{Mid-sequence forgetting.}
The fixed-size memory provides global context (V-Sink) and local context (sliding window) but has no mechanism for recalling content generated between the two; mid-sequence details are progressively lost at each KV Flush (see supplementary).

\begin{figure*}[t]
\centering
\setlength{\tabcolsep}{2pt} 
\begin{tabular}{rccccc}
& \scriptsize $t=0\text{s}$ & \scriptsize $t=60\text{s}$ & \scriptsize $t=120\text{s}$ & \scriptsize $t=180\text{s}$ & \scriptsize $t=240\text{s}$ \\ [0.1em]

\rotatebox{90}{\hspace{-3em}\scriptsize River} &
\bmvaHangBox{\includegraphics[width=0.18\linewidth]{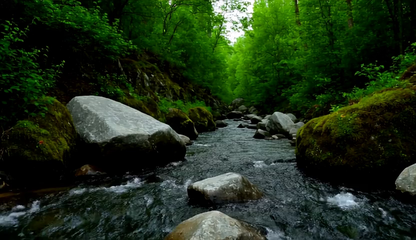}} &
\bmvaHangBox{\includegraphics[width=0.18\linewidth]{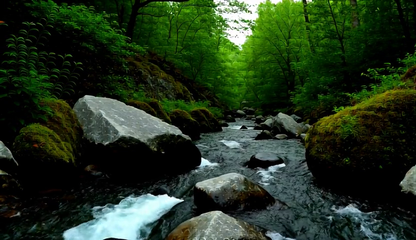}} &
\bmvaHangBox{\includegraphics[width=0.18\linewidth]{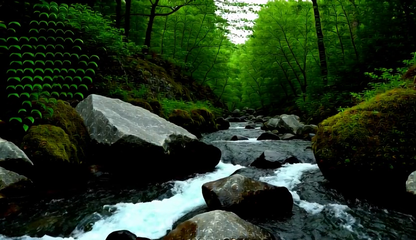}} &
\bmvaHangBox{\includegraphics[width=0.18\linewidth]{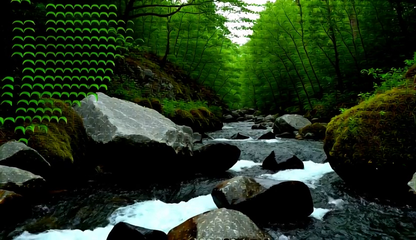}} &
\bmvaHangBox{\includegraphics[width=0.18\linewidth]{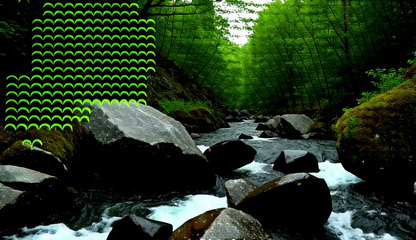}}
\vspace{0mm} \\

\rotatebox{90}{\hspace{-3em}\scriptsize Rain} &
\bmvaHangBox{\includegraphics[width=0.18\linewidth]{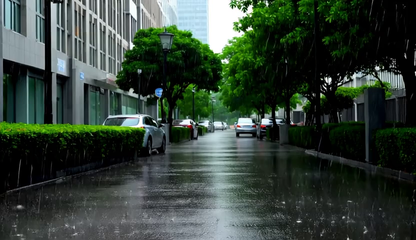}} &
\bmvaHangBox{\includegraphics[width=0.18\linewidth]{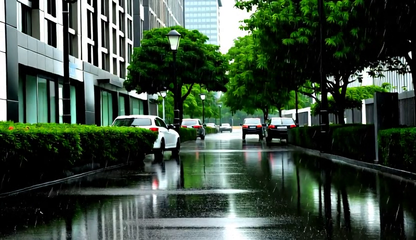}} &
\bmvaHangBox{\includegraphics[width=0.18\linewidth]{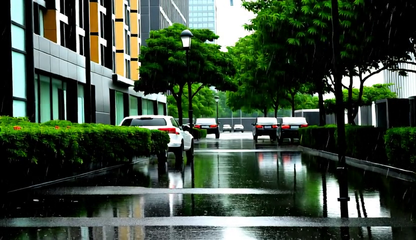}} &
\bmvaHangBox{\includegraphics[width=0.18\linewidth]{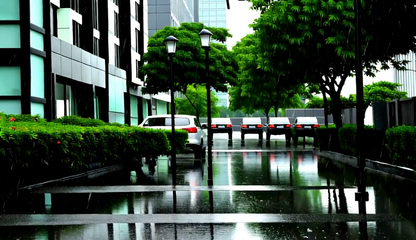}} &
\bmvaHangBox{\includegraphics[width=0.18\linewidth]{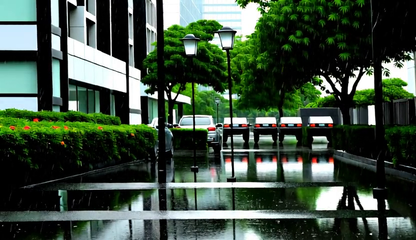}}

\end{tabular}
\caption{\textbf{Limitation analysis on ultra-long horizons (240s).}
Steady-Forcing remains structurally coherent in these examples, but extreme rollouts reveal residual limitations, including loss of fine high-frequency background texture consistency, mild color drift, and partial motion stagnation in large flow regions.}
\label{fig:limitation_results}
\end{figure*}

\textbf{Residual stagnation.}
Motion stagnation in large-area flow regions (open ocean, wide rivers) remains partially unresolved at longer durations
(Fig.~\ref{fig:limitation_results}); the EMA-Sink's single compressed state cannot represent broad-scale coherent dynamics, and the flush period is not adaptive to local motion complexity.

\textbf{Future work.}
Replacing the fixed-size EMA-Sink with hierarchical memory~\cite{mamba,mamba2}, scaling to stronger teachers~\cite{wan2025, helios, li2025stable} via VLM-integrated Re-DMD~\cite{lu2025reward}, and adopting adaptive flush scheduling are promising directions for addressing the above limitations.

\section{Conclusion}

We presented Steady-Forcing, a training and inference framework that addresses the stability--motion trade-off in long-horizon fixed-camera nature video generation. By combining a persistent V-Sink for background identity, an EMA-Sink for compressed motion memory, Block-Relativistic RoPE for extended temporal encoding, and Periodic KV Flush for cache purification, Steady-Forcing improves the balance between background stability and fluid dynamics over multi-minute autoregressive rollouts. Motion-rewarded prior initialization and Self-Forcing distillation from a Wan2.1-14B teacher with domain-specific negative prompting specialize the generator for fixed-camera natural flow without ground-truth video supervision. Evaluations across seven baselines on VBench and a blind user study show improvements in long-horizon background consistency, imaging quality, and perceived motion continuity. The quantitative evaluation further suggests that aggregate VBench scores under-penalize fixed-camera artifacts such as color drift, texture hardening and flow stagnation while rewarding camera drift-induced optical flow as Dynamic Degree, establishing a clear need for task-specific evaluation protocols for static-camera nature-stream generation.


\bibliography{egbib}


\clearpage 

\appendix
\setcounter{page}{1}       
\setcounter{section}{0}    
\setcounter{figure}{0}     
\setcounter{table}{0}      

\renewcommand{\thefigure}{S\arabic{figure}}
\renewcommand{\thetable}{S\arabic{table}}
\renewcommand{\thepage}{S\arabic{page}}



\renewcommand{\thefigure}{S\arabic{figure}}
\renewcommand{\thetable}{S\arabic{table}}
\renewcommand{\theequation}{S\arabic{equation}}

\renewcommand{\thesection}{\Alph{section}}
\renewcommand{\thesubsection}{\Alph{section}.\arabic{subsection}}

\def\eg{\emph{e.g}\bmvaOneDot}
\def\Eg{\emph{E.g}\bmvaOneDot}
\def\etal{\emph{et al}\bmvaOneDot}
\def\etc{\emph{etc}\bmvaOneDot}

\newcommand{\sfm}{\emph{Steady-Forcing}}
\newcommand{\vsink}{\textbf{V-Sink}}
\newcommand{\emasink}{\textbf{EMA-Sink}}
\newcommand{\kvflush}{\textbf{Periodic KV Flush}}
\newcommand{\Sfixed}{\ensuremath{S_{\text{fixed}}}}

\title{Supplementary Material for Steady-Forcing: Balancing Spatial Persistence and Motion Continuity in Long-Horizon Nature Video Diffusion}

\runninghead{Steady-Forcing}{Supplementary Material}

\begin{center}
    \vspace*{1cm}
    {\LARGE \bf Supplementary Material: \\ Steady-Forcing: Balancing Spatial Persistence and Motion Continuity in Long-Horizon Nature Video Diffusion \par}
    \vspace*{1cm}
    \hrule 
\end{center}
\vspace*{0.5cm}

\maketitle

\noindent
This supplementary material accompanies the main paper.
Figures, tables, and equations are numbered with the prefix ``S''
(\eg, Fig.~S1, Table~S1, Eq.~S1) to distinguish them from
main-paper elements. We refer to main-paper figures and tables as \textbf{Main Fig.~X} and \textbf{Main Table~X}. We make code, data, model, and generated videos public at: \url{https://minar09.github.io/steadyforcing/}

\section{Additional Qualitative Results}
\label{sec:supp_qual}

\subsection{Extended Rollouts Across Scene Categories}
\label{sec:supp_qual_extended}

\begin{figure*}[t]
\centering
\setlength{\tabcolsep}{1pt}
\begin{tabular}{@{}c@{\hspace{2pt}}cccc@{}}
& \scriptsize $t=0\text{s}$ & \scriptsize $t=20\text{s}$
& \scriptsize $t=40\text{s}$ & \scriptsize $t=60\text{s}$ \\
\rotatebox{90}{\hspace{-2em}\small Fire} &
\bmvaHangBox{\includegraphics[width=0.24\linewidth]{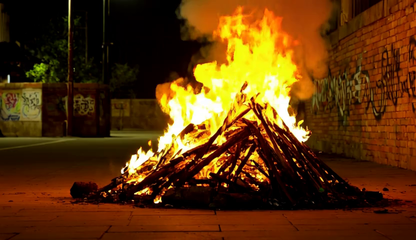}} &
\bmvaHangBox{\includegraphics[width=0.24\linewidth]{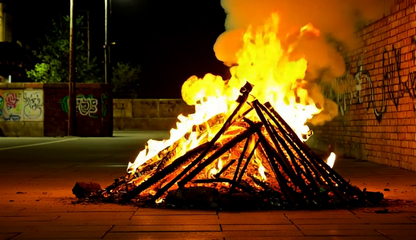}} &
\bmvaHangBox{\includegraphics[width=0.24\linewidth]{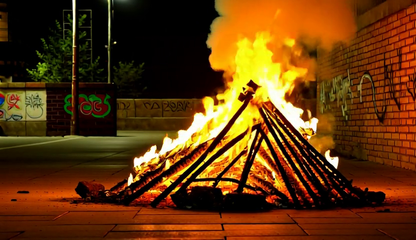}} &
\bmvaHangBox{\includegraphics[width=0.24\linewidth]{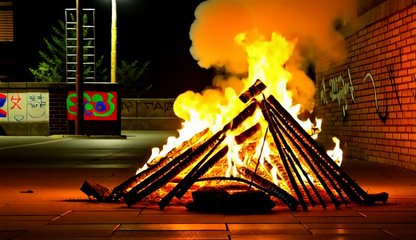}} \\
\vspace{-5mm} \\
\rotatebox{90}{\hspace{-2em}\small Flood} &
\bmvaHangBox{\includegraphics[width=0.24\linewidth]{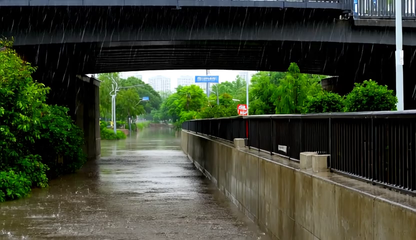}} &
\bmvaHangBox{\includegraphics[width=0.24\linewidth]{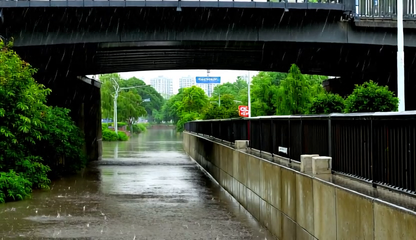}} &
\bmvaHangBox{\includegraphics[width=0.24\linewidth]{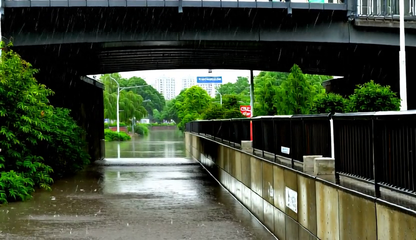}} &
\bmvaHangBox{\includegraphics[width=0.24\linewidth]{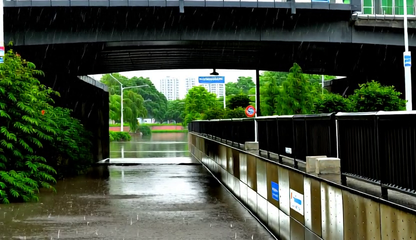}} \\
\vspace{-5mm} \\
\rotatebox{90}{\hspace{-2em}\small Forest} &
\bmvaHangBox{\includegraphics[width=0.24\linewidth]{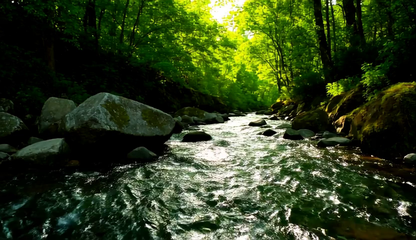}} &
\bmvaHangBox{\includegraphics[width=0.24\linewidth]{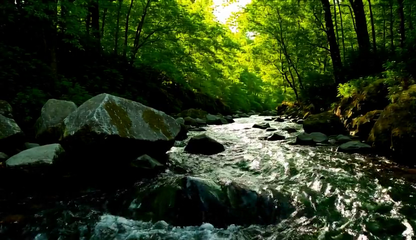}} &
\bmvaHangBox{\includegraphics[width=0.24\linewidth]{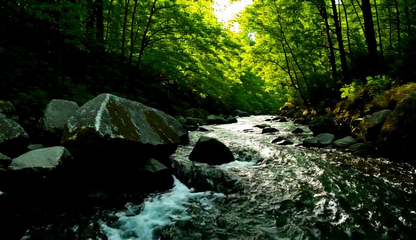}} &
\bmvaHangBox{\includegraphics[width=0.24\linewidth]{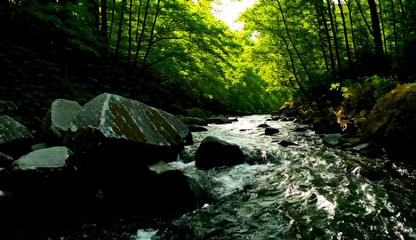}} \\
\vspace{-5mm} \\
\rotatebox{90}{\hspace{-2em}\small Lava} &
\bmvaHangBox{\includegraphics[width=0.24\linewidth]{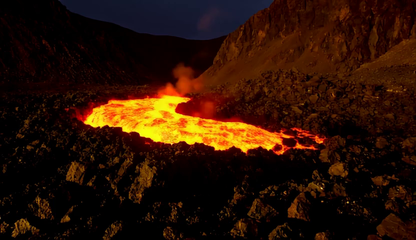}} &
\bmvaHangBox{\includegraphics[width=0.24\linewidth]{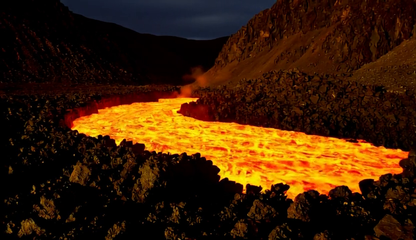}} &
\bmvaHangBox{\includegraphics[width=0.24\linewidth]{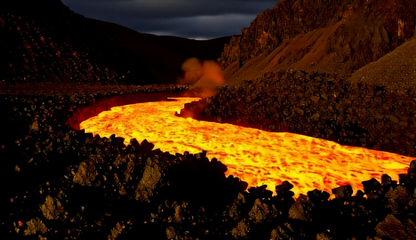}} &
\bmvaHangBox{\includegraphics[width=0.24\linewidth]{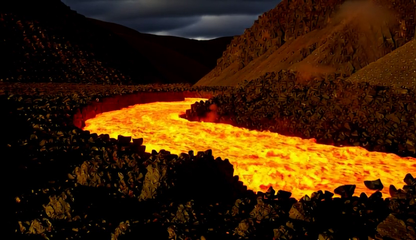}} \\
\vspace{-5mm} \\
\rotatebox{90}{\hspace{-2em}\small Rain} &
\bmvaHangBox{\includegraphics[width=0.24\linewidth]{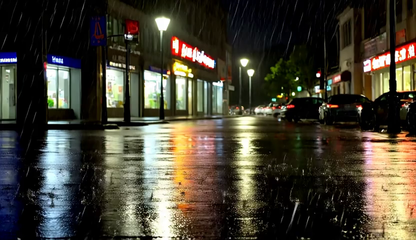}} &
\bmvaHangBox{\includegraphics[width=0.24\linewidth]{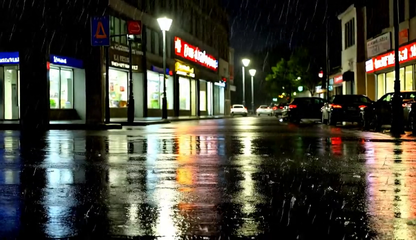}} &
\bmvaHangBox{\includegraphics[width=0.24\linewidth]{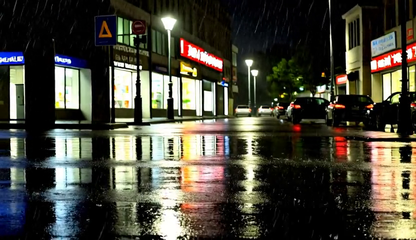}} &
\bmvaHangBox{\includegraphics[width=0.24\linewidth]{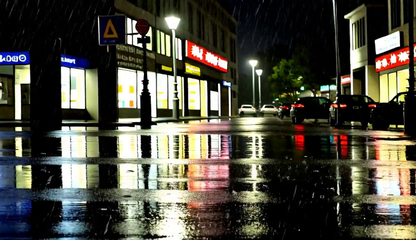}} \\
\vspace{-5mm} \\
\rotatebox{90}{\hspace{-2em}\small Hill} &
\bmvaHangBox{\includegraphics[width=0.24\linewidth]{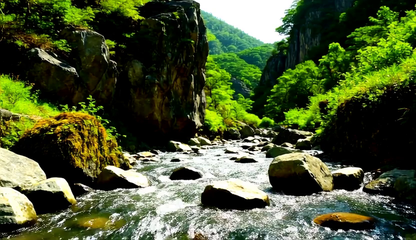}} &
\bmvaHangBox{\includegraphics[width=0.24\linewidth]{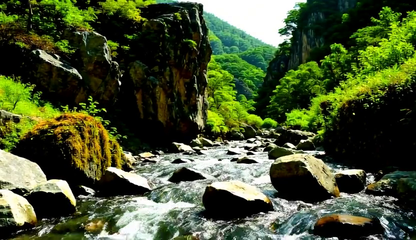}} &
\bmvaHangBox{\includegraphics[width=0.24\linewidth]{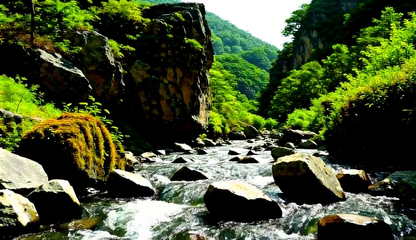}} &
\bmvaHangBox{\includegraphics[width=0.24\linewidth]{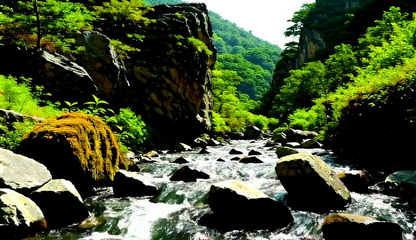}} \\
\end{tabular}
\caption{\textbf{Extended rollouts across six scene categories.}
Each row shows a single-prompt continuous generation at
$t \in \{0,20,40,60\}$\,s. \sfm{} preserves background layout and
category-specific fluid motion throughout each rollout.}
\label{fig:supp_categories}
\end{figure*}

Figure~\ref{fig:supp_categories} shows 4-frame strips at
$t \in \{0,20,40,60\}$\,s for six fixed-camera nature-flow
categories, each generated in a single continuous rollout by
\sfm{}.
The \vsink{} maintains stable background identity across all
categories, while the \emasink{} preserves category-specific
motion: directional flow in rivers, upward drift in smoke,
wave turbulence in ocean, and particle coherence in rain and snow.

\subsection{Extended Baseline Comparison (Additional Prompts)}
\label{sec:supp_qual_comparison}

\begin{figure*}[p] 
\centering
\setlength{\tabcolsep}{1pt} 
\begin{tabular}{rccc}
& {\scriptsize $t=0\text{s}$} & {\scriptsize $t=30\text{s}$} & {\scriptsize $t=60\text{s}$} \\[0em]

\parbox[c]{2cm}{\raggedleft\scriptsize $^\dagger$CausVid \cite{yin2025causvid}} &
\bmvaHangBox{\includegraphics[width=0.28\linewidth]{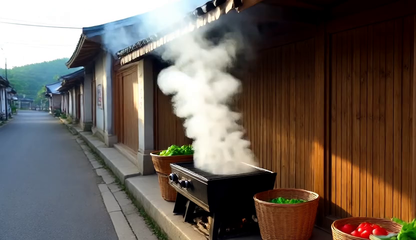}} &
\bmvaHangBox{\includegraphics[width=0.28\linewidth]{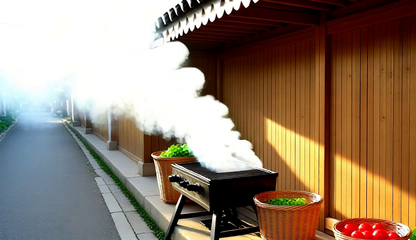}} &
\bmvaHangBox{\includegraphics[width=0.28\linewidth]{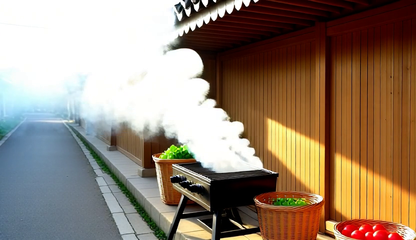}} \\ [0em]

\parbox[c]{2cm}{\raggedleft\scriptsize $^\dagger$Self-\\Forcing \cite{huang2025selfforcing}} &
\bmvaHangBox{\includegraphics[width=0.28\linewidth]{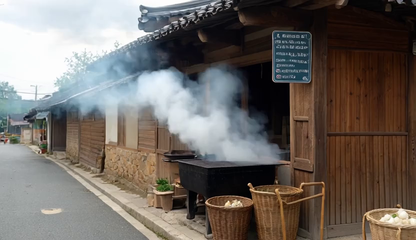}} &
\bmvaHangBox{\includegraphics[width=0.28\linewidth]{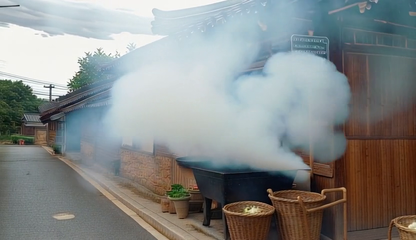}} &
\bmvaHangBox{\includegraphics[width=0.28\linewidth]{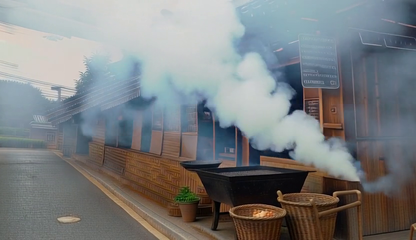}} \\ [0em]

\parbox[c]{2cm}{\raggedleft\scriptsize $^\dagger$Infinite-\\Forcing \cite{infinite-forcing}} &
\bmvaHangBox{\includegraphics[width=0.28\linewidth]{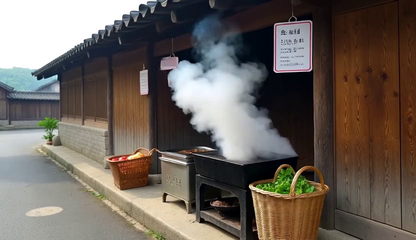}} &
\bmvaHangBox{\includegraphics[width=0.28\linewidth]{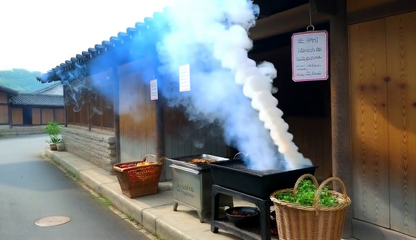}} &
\bmvaHangBox{\includegraphics[width=0.28\linewidth]{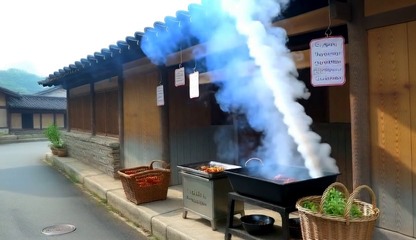}} \\ [0em]

\parbox[c]{2cm}{\raggedleft\scriptsize $^\dagger$Rolling-\\Forcing \cite{liu2025rolling}} &
\bmvaHangBox{\includegraphics[width=0.28\linewidth]{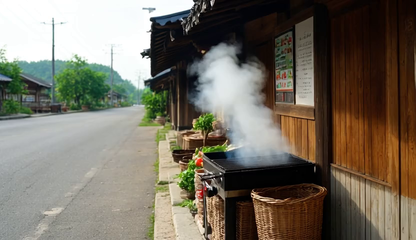}} &
\bmvaHangBox{\includegraphics[width=0.28\linewidth]{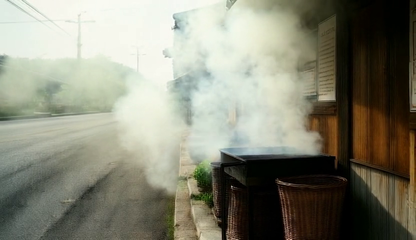}} &
\bmvaHangBox{\includegraphics[width=0.28\linewidth]{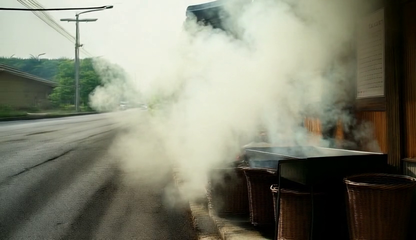}} \\ [0em]

\parbox[c]{2cm}{\raggedleft\scriptsize $^\dagger$Reward-\\Forcing \cite{lu2025reward}} &
\bmvaHangBox{\includegraphics[width=0.28\linewidth]{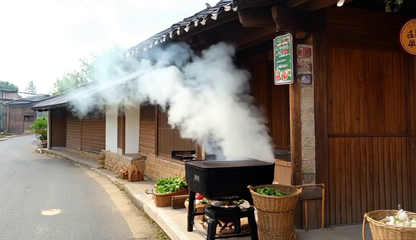}} &
\bmvaHangBox{\includegraphics[width=0.28\linewidth]{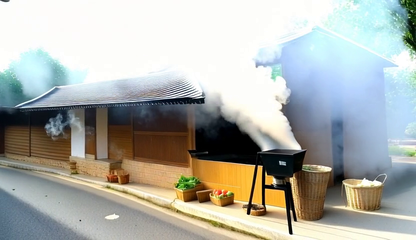}} &
\bmvaHangBox{\includegraphics[width=0.28\linewidth]{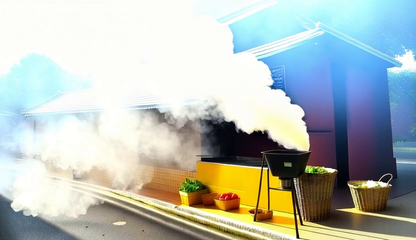}} \\ [0em]

\parbox[c]{2cm}{\raggedleft\scriptsize $^\dagger$LongLive \cite{yang2025longlive}} &
\bmvaHangBox{\includegraphics[width=0.28\linewidth]{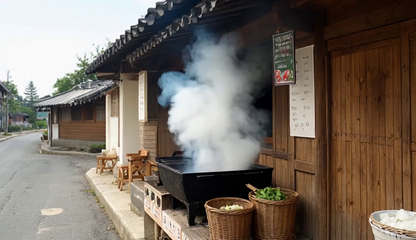}} &
\bmvaHangBox{\includegraphics[width=0.28\linewidth]{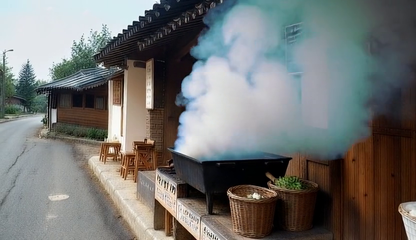}} &
\bmvaHangBox{\includegraphics[width=0.28\linewidth]{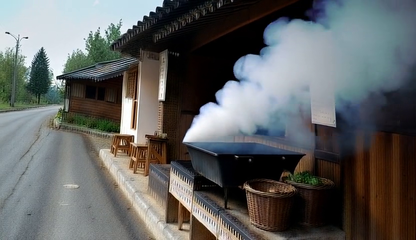}} \\ [0.3em]

\parbox[c]{2cm}{\raggedleft\scriptsize $^\dagger$Causal-\\Forcing \cite{zhu2026causal}} &
\bmvaHangBox{\includegraphics[width=0.28\linewidth]{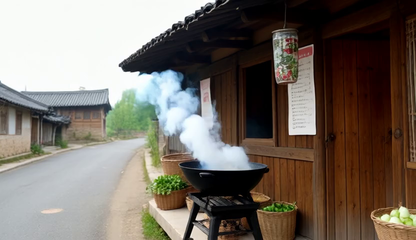}} &
\bmvaHangBox{\includegraphics[width=0.28\linewidth]{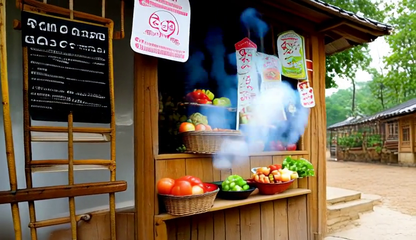}} &
\bmvaHangBox{\includegraphics[width=0.28\linewidth]{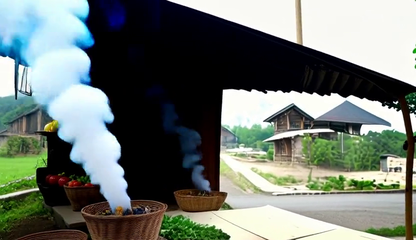}} \\ [0.3em]

\parbox[c]{2cm}{\raggedleft\scriptsize \textbf{Steady-}\\ \textbf{Forcing (Ours)}} &
\bmvaHangBox{\includegraphics[width=0.28\linewidth]{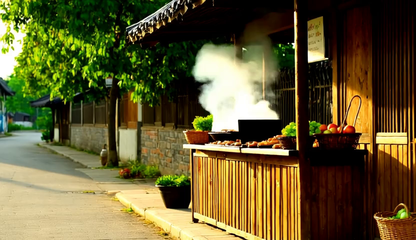}} &
\bmvaHangBox{\includegraphics[width=0.28\linewidth]{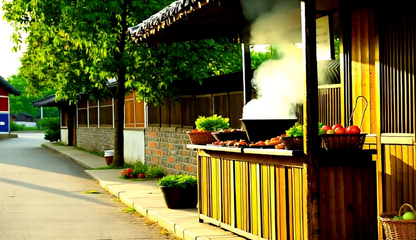}} &
\bmvaHangBox{\includegraphics[width=0.28\linewidth]{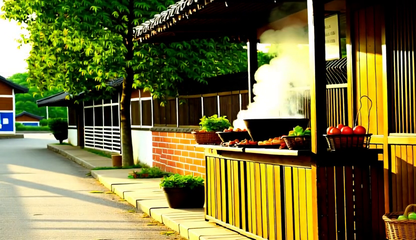}} \\

\end{tabular}
\caption{\textbf{Extended baseline comparison at 60 seconds.}
Each row shows one method; each column shows a sampled timestep.
$^\dagger$~Public baseline weights evaluated under our fixed-camera inference protocol.}
\label{fig:supp_comparison}
\end{figure*}

Figure~\ref{fig:supp_comparison} extends \textbf{Main Fig.~5} to
additional scene types using the same task-adapted protocol
($\dagger$).

\subsection{Failure Cases}
\label{sec:supp_failure}

\begin{figure*}[p] 
\centering
\setlength{\tabcolsep}{1pt} 
\begin{tabular}{ccc}
& {\small Ocean Scene} & {\small Smoke Scene} \\[0em]

\parbox[c]{1cm}{\raggedleft\small $t=0\text{s}$} &
\bmvaHangBox{\includegraphics[width=0.44\linewidth]{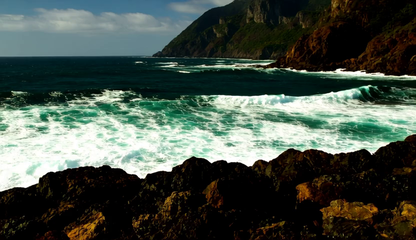}} &
\bmvaHangBox{\includegraphics[width=0.44\linewidth]{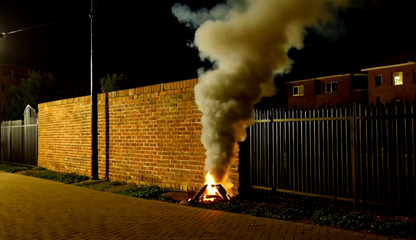}} \\ [0.2em]

\parbox[c]{1cm}{\raggedleft\small $t=60\text{s}$} &
\bmvaHangBox{\includegraphics[width=0.44\linewidth]{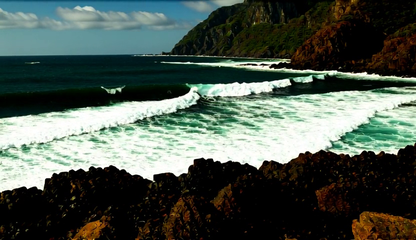}} &
\bmvaHangBox{\includegraphics[width=0.44\linewidth]{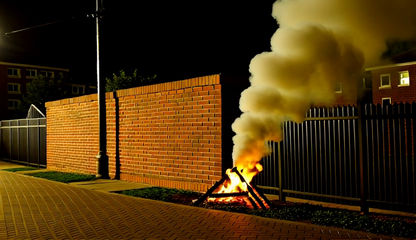}} \\ [0.2em]

\parbox[c]{1cm}{\raggedleft\small $t=120\text{s}$} &
\bmvaHangBox{\includegraphics[width=0.44\linewidth]{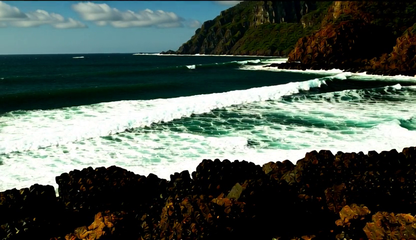}} &
\bmvaHangBox{\includegraphics[width=0.44\linewidth]{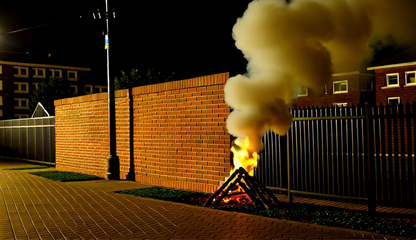}} \\ [0.2em]

\parbox[c]{1cm}{\raggedleft\small $t=180\text{s}$} &
\bmvaHangBox{\includegraphics[width=0.44\linewidth]{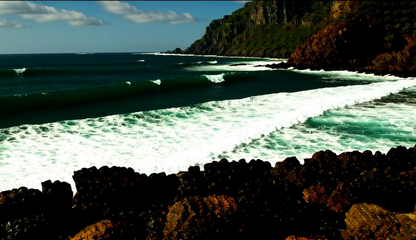}} &
\bmvaHangBox{\includegraphics[width=0.44\linewidth]{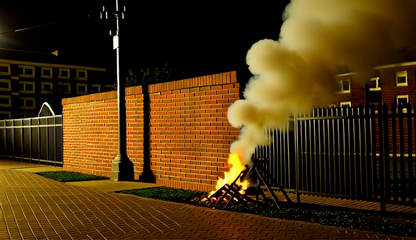}} \\ [0.2em]

\parbox[c]{1cm}{\raggedleft\small $t=240\text{s}$} &
\bmvaHangBox{\includegraphics[width=0.44\linewidth]{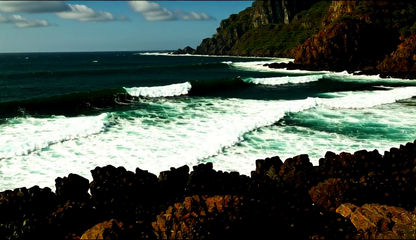}} &
\bmvaHangBox{\includegraphics[width=0.44\linewidth]{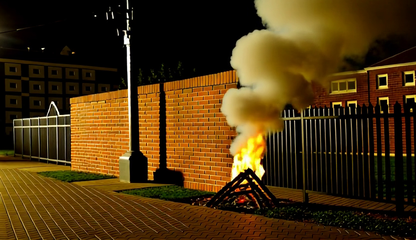}} \\

\end{tabular}
\caption{\textbf{Vertical sequence analysis of long-horizon failure cases (120s--240s).} 
Left column: motion flattening artifacts appearing within the open-ocean environment over extended time. Right column: dense particle distribution drift and blurring in the dense smoke scene. Both failure modes underscore generation boundary constraints tied to the base generation model's training data distribution limits.}
\label{fig:supp_failure}
\end{figure*}

Figure~\ref{fig:supp_failure} illustrates the two primary failure modes
of \sfm{} at extreme horizons ($t \geq 120$\,s):

\begin{itemize}
  \item \textbf{Motion flattening.} In open-ocean scenes, wave amplitude
    decays progressively as the \emasink{}'s single compressed state
    cannot represent broad-scale coherent dynamics.
  \item \textbf{Texture hardening.} Rare cache contamination events
    that survive a flush can reinforce subtle repeated patterns in
    textured static regions over time.
  \item \textbf{Temporal repetition.} In low-optical-flow scenes
    (\eg, still water), the model occasionally enters a short-period
    loop where 2--3 frames cycle.
  \item \textbf{Local geometric drift.} Dense particle effects
    (heavy rain, thick smoke) can cause minor background displacement
    at scene boundaries over very long rollouts.
\end{itemize}


\section{Additional Quantitative Results}
\label{sec:supp_quant}

\subsection{Full VBench Tables}
\label{sec:supp_quant_full}

Tables~\ref{tab:supp_5s}--\ref{tab:supp_240s} report all six evaluated
VBench~\cite{huang2023vbench} dimensions for every method at 4 horizons, i.e., 5s, 60s, 120s and 240s tiers.

\begin{table*}[t]
\centering
\scriptsize
\setlength{\tabcolsep}{5pt} 
\begin{tabular}{lccccccc}
\hline
Method & 
\begin{tabular}[c]{@{}c@{}}Background\\Consistency $\uparrow$\end{tabular} & 
\begin{tabular}[c]{@{}c@{}}Motion\\Smoothness $\uparrow$\end{tabular} & 
\begin{tabular}[c]{@{}c@{}}Imaging\\Quality $\uparrow$\end{tabular} & 
\begin{tabular}[c]{@{}c@{}}Temporal\\Flickering $\uparrow$\end{tabular} & 
\begin{tabular}[c]{@{}c@{}}Aesthetic\\Quality $\uparrow$\end{tabular} & 
\begin{tabular}[c]{@{}c@{}}Dynamic\\Degree $\uparrow$\end{tabular} & 
\begin{tabular}[c]{@{}c@{}}Horizon\\Average $\uparrow$\end{tabular} \\ \hline
$^\dagger$CausVid \cite{yin2025causvid}            & 96.49 & 98.65 & 51.79 & 97.99 & 61.06 & 37.01 & 73.83 \\
$^\dagger$Self-Forcing \cite{huang2025selfforcing} & 96.47 & 98.86 & 62.50 & 98.30 & 57.58 & \underline{48.02} & 76.96 \\
$^\dagger$Infinite-Forcing \cite{infinite-forcing} & \underline{97.46} & \underline{98.96} & 63.21 & \textbf{98.68} & 56.37 & 25.70 & 73.40 \\
$^\dagger$Rolling-Forcing \cite{liu2025rolling}    & 96.63 & 98.55 & 64.69 & 97.65 & 60.81 & 47.06 & \underline{77.57} \\
$^\dagger$Reward-Forcing \cite{lu2025reward}       & 96.62 & 98.67 & 65.55 & 98.20 & 58.46 & 27.95 & 74.24 \\
$^\dagger$LongLive \cite{yang2025longlive}         & 95.88 & 98.72 & \underline{67.13} & 98.07 & 57.23 & 34.26 & 75.22 \\
$^\dagger$Causal-Forcing \cite{zhu2026causal}      & 95.85 & 97.95 & 63.46 & 96.16 & \underline{62.05} & \textbf{70.00} & \textbf{80.91} \\
\textbf{Steady-Forcing (Ours)}           & \textbf{98.06} & \textbf{99.00} & \textbf{69.87} & \underline{98.53} & \textbf{63.50} & 25.22 & 75.70 \\ \hline
\end{tabular}
\vspace{0.1em}
\caption{\textbf{Full VBench quantitative results at the 5-second horizon (Tier 1).} We show complete per-criterion scores. $^\dagger$ indicates public baseline execution under our uniform fixed-camera evaluation environment.}
\label{tab:supp_5s}
\end{table*}


\begin{table*}[t]
\centering
\scriptsize
\setlength{\tabcolsep}{5pt} 
\begin{tabular}{lccccccc}
\hline
Method & 
\begin{tabular}[c]{@{}c@{}}Background\\Consistency $\uparrow$\end{tabular} & 
\begin{tabular}[c]{@{}c@{}}Motion\\Smoothness $\uparrow$\end{tabular} & 
\begin{tabular}[c]{@{}c@{}}Imaging\\Quality $\uparrow$\end{tabular} & 
\begin{tabular}[c]{@{}c@{}}Temporal\\Flickering $\uparrow$\end{tabular} & 
\begin{tabular}[c]{@{}c@{}}Aesthetic\\Quality $\uparrow$\end{tabular} & 
\begin{tabular}[c]{@{}c@{}}Dynamic\\Degree $\uparrow$\end{tabular} & 
\begin{tabular}[c]{@{}c@{}}Horizon\\Average $\uparrow$\end{tabular} \\ \hline
$^\dagger$CausVid \cite{yin2025causvid}            & 92.51 & 98.99 & 55.72 & 98.30 & \underline{63.51} & 40.93 & 74.99 \\
$^\dagger$Self-Forcing \cite{huang2025selfforcing} & 90.55 & 98.73 & 62.26 & 98.03 & 58.11 & \underline{65.88} & \underline{78.93} \\
$^\dagger$Infinite-Forcing \cite{infinite-forcing} & \underline{92.66} & \textbf{99.16} & 56.14 & \textbf{98.88} & 59.98 & 26.46 & 72.21 \\
$^\dagger$Rolling-Forcing \cite{liu2025rolling}    & 92.05 & 98.63 & 59.84 & 97.89 & 58.61 & 48.89 & 75.99 \\
$^\dagger$Reward-Forcing \cite{lu2025reward}       & 90.62 & 98.65 & 63.29 & 97.94 & 59.58 & 43.61 & 75.62 \\
$^\dagger$LongLive \cite{yang2025longlive}         & 90.83 & 98.44 & 61.45 & 97.67 & 57.05 & 59.18 & 77.44 \\
$^\dagger$Causal-Forcing \cite{zhu2026causal}      & 87.73 & 97.05 & \underline{68.30} & 94.04 & 55.91 & \textbf{91.37} & \textbf{82.40} \\
\textbf{Steady-Forcing (Ours)}           & \textbf{95.60} & \underline{99.13} & \textbf{71.59} & \underline{98.44} & \textbf{64.85} & 37.43 & 77.84 \\ \hline
\end{tabular}
\vspace{0.1em}
\caption{\textbf{Full VBench quantitative results at the 60-second horizon (Tier 2).} We show complete per-criterion scores. $^\dagger$ indicates public baseline execution under our uniform fixed-camera evaluation environment.}
\label{tab:supp_60s}
\end{table*}


\begin{table*}[t]
\centering
\scriptsize
\setlength{\tabcolsep}{5pt} 
\begin{tabular}{lccccccc}
\hline
Method & 
\begin{tabular}[c]{@{}c@{}}Background\\Consistency $\uparrow$\end{tabular} & 
\begin{tabular}[c]{@{}c@{}}Motion\\Smoothness $\uparrow$\end{tabular} & 
\begin{tabular}[c]{@{}c@{}}Imaging\\Quality $\uparrow$\end{tabular} & 
\begin{tabular}[c]{@{}c@{}}Temporal\\Flickering $\uparrow$\end{tabular} & 
\begin{tabular}[c]{@{}c@{}}Aesthetic\\Quality $\uparrow$\end{tabular} & 
\begin{tabular}[c]{@{}c@{}}Dynamic\\Degree $\uparrow$\end{tabular} & 
\begin{tabular}[c]{@{}c@{}}Horizon\\Average $\uparrow$\end{tabular} \\ \hline
$^\dagger$CausVid \cite{yin2025causvid}            & \underline{92.75} & \underline{98.97} & 56.95 & \underline{98.64} & \underline{62.06} & 20.47 & 71.64 \\
$^\dagger$Self-Forcing \cite{huang2025selfforcing} & 88.55 & 98.86 & 57.66 & 98.45 & 51.48 & 29.93 & 70.82 \\
$^\dagger$Infinite-Forcing \cite{infinite-forcing} & 91.13 & \textbf{99.22} & 48.00 & \textbf{98.99} & 58.18 & 24.97 & 70.08 \\
$^\dagger$Rolling-Forcing \cite{liu2025rolling}    & 89.75 & 98.77 & 55.26 & 98.13 & 53.98 & 42.30 & 73.03 \\
$^\dagger$Reward-Forcing \cite{lu2025reward}       & 87.65 & 98.63 & 52.63 & 97.70 & 54.44 & 44.96 & 72.67 \\
$^\dagger$LongLive \cite{yang2025longlive}         & 88.50 & 98.48 & 53.32 & 97.74 & 52.80 & \underline{52.25} & 73.85 \\
$^\dagger$Causal-Forcing \cite{zhu2026causal}      & 88.09 & 96.68 & \underline{59.91} & 93.43 & 56.33 & \textbf{95.95} & \textbf{81.73} \\
\textbf{Steady-Forcing (Ours)}           & \textbf{95.07} & \textbf{99.22} & \textbf{66.21} & \underline{98.64} & \textbf{62.16} & 37.99 & \underline{76.55} \\ \hline
\end{tabular}
\vspace{0.1em}
\caption{\textbf{Full VBench quantitative results at the 120-second horizon (Tier 3).} We show complete per-criterion scores. $^\dagger$ indicates public baseline execution under our uniform fixed-camera evaluation environment.}
\label{tab:supp_120s}
\end{table*}


\begin{table*}[t]
\centering
\scriptsize
\setlength{\tabcolsep}{5pt} 
\begin{tabular}{lccccccc}
\hline
Method & 
\begin{tabular}[c]{@{}c@{}}Background\\Consistency $\uparrow$\end{tabular} & 
\begin{tabular}[c]{@{}c@{}}Motion\\Smoothness $\uparrow$\end{tabular} & 
\begin{tabular}[c]{@{}c@{}}Imaging\\Quality $\uparrow$\end{tabular} & 
\begin{tabular}[c]{@{}c@{}}Temporal\\Flickering $\uparrow$\end{tabular} & 
\begin{tabular}[c]{@{}c@{}}Aesthetic\\Quality $\uparrow$\end{tabular} & 
\begin{tabular}[c]{@{}c@{}}Dynamic\\Degree $\uparrow$\end{tabular} & 
\begin{tabular}[c]{@{}c@{}}Horizon\\Average $\uparrow$\end{tabular} \\ \hline
$^\dagger$CausVid \cite{yin2025causvid}            & 88.25 & 98.69 & 64.27 & 98.43 & \textbf{61.51} & 23.48 & 72.44 \\
$^\dagger$Self-Forcing \cite{huang2025selfforcing} & 83.70 & 98.54 & 55.73 & 98.01 & 49.70 & 46.89 & 72.10 \\
$^\dagger$Infinite-Forcing \cite{infinite-forcing} & 85.95 & \textbf{99.37} & 54.92 & \textbf{99.24} & 55.38 & 26.62 & 70.24 \\
$^\dagger$Rolling-Forcing \cite{liu2025rolling}    & 87.37 & 99.07 & 57.29 & 98.37 & 51.80 & \underline{55.88} & 74.96 \\
$^\dagger$Reward-Forcing \cite{lu2025reward}       & 86.17 & 98.51 & 61.25 & 97.85 & 53.16 & 52.93 & \underline{74.98} \\
$^\dagger$LongLive \cite{yang2025longlive}         & 84.58 & 98.69 & 55.95 & 97.85 & 46.83 & 35.80 & 69.95 \\
$^\dagger$Causal-Forcing \cite{zhu2026causal}      & 86.87 & 96.47 & \underline{66.36} & 92.37 & 54.79 & \textbf{94.75} & \textbf{81.94} \\
\textbf{Steady-Forcing (Ours)}                     & \textbf{92.57} & \underline{99.34} & \textbf{71.27} & \underline{99.14} & \underline{60.76} & 21.86 & 74.16 \\ \hline
\end{tabular}
\vspace{0.1em}
\caption{\textbf{Full VBench quantitative results at the ultra-long 240-second horizon (Tier 4).} We present complete, per-criterion scores for all baseline methods. $^\dagger$ indicates public baseline execution under our uniform fixed-camera evaluation environment.}
\label{tab:supp_240s}
\end{table*}

\subsection{Native-Configuration Baseline Results}
\label{sec:supp_native}

The main paper evaluates baselines under our task-adapted inference
protocol ($\dagger$) for a fair comparison.

\subsection{Analysis: Dynamic Degree and the quantitative Evaluation}
\label{sec:supp_dd_analysis}

The quantitative results in Tables~\ref{tab:supp_5s},~\ref{tab:supp_60s},~\ref{tab:supp_120s},~\ref{tab:supp_240s} require careful interpretation because the VBench~\cite{huang2023vbench} aggregate average does not directly measure the qualities most critical for fixed-camera static-scene generation.

\textbf{Why Steady-Forcing's Dynamic Degree is low.}
Dynamic Degree is computed from mean optical flow magnitude across
the full frame. In a fixed-camera setting, two physically distinct
sources of optical flow both contribute to this score: (a)~genuine
fluid motion in dynamic foreground regions (rivers, fire, smoke, etc.),
and (b)~background drift caused by exposure bias. A method that
successfully suppresses camera drift and background displacement
necessarily reduces total frame-level optical flow — and VBench
registers this as lower Dynamic Degree, regardless of whether actual
fluid motion is preserved. Steady-Forcing's Background Consistency stays at the highest among all evaluated methods across all horizons by a
substantial margin — confirms that its lower Dynamic Degree reflects
stable background anchoring rather than fluid motion collapse.

\textbf{Why Causal-Forcing's Dynamic Degree is high.}
Across all horizons, Causal-Forcing reports the highest Dynamic
Degree and VBench average, yet its Background Consistency is
consistently the lowest among all methods across horizons: 5s (95.85), 60s (87.73) and 120s (88.09), declining further at 240s (86.87); using their public chunkwise weights under our inference protocol for static-view continuous-flow. This inverse correlation — highest motion score, lowest spatial stability — is precisely the pattern produced by camera drift: rotating or translating backgrounds generate large optical flow vectors that VBench rewards as high Dynamic Degree, while the rigid spatial identity of the scene degrades. Under our task-adapted chunk-wise evaluation, Causal-Forcing exhibits progressive scene rotation that inflates Dynamic Degree without preserving static background identity. This confirms that VBench Dynamic Degree cannot reliably distinguish desired fluid motion from undesired camera drift in a fixed-camera setting.

\textbf{The case for a task-specific benchmark.}
Table~\ref{tab:supp_dd_comparison} summarizes the relationship
between Dynamic Degree and Background Consistency across horizons,
illustrating that a method can rank first on VBench average while
exhibiting the worst background stability — a direct contradiction
of the fixed-camera nature-stream task requirement. Conversely,
Steady-Forcing ranks last or second-last in Dynamic Degree at every
horizon while ranking first in Background Consistency and Imaging
Quality. This systematic misalignment demonstrates that aggregate
VBench scoring is insufficient for our task: it does not penalize
texture hardening, flow stagnation, or background drift in
static-camera scenes, and it rewards drift-induced optical flow as
a proxy for desired motion. These observations directly motivate a
future task-specific evaluation protocol — such as the
Steady Nature Flow benchmark concept introduced in earlier iterations
of this work — that jointly measures background geometric stability,
flow persistence over time, and artifact suppression as independent
criteria rather than folding them into a single aggregate score.

\begin{table*}[t]
\centering
\scriptsize
\setlength{\tabcolsep}{4pt} 
\begin{tabular}{lcccccc}
\hline
 & \multicolumn{2}{c}{60s Horizon} & \multicolumn{2}{c}{120s Horizon} & \multicolumn{2}{c}{240s Horizon} \\
\cline{2-7}
Method & 
\begin{tabular}[c]{@{}c@{}}Background\\Consistency $\uparrow$\end{tabular} & 
\begin{tabular}[c]{@{}c@{}}Dynamic\\Degree $\uparrow$\end{tabular} & 
\begin{tabular}[c]{@{}c@{}}Background\\Consistency $\uparrow$\end{tabular} & 
\begin{tabular}[c]{@{}c@{}}Dynamic\\Degree $\uparrow$\end{tabular} & 
\begin{tabular}[c]{@{}c@{}}Background\\Consistency $\uparrow$\end{tabular} & 
\begin{tabular}[c]{@{}c@{}}Dynamic\\Degree $\uparrow$\end{tabular} \\
\hline
$^\dagger$Causal-Forcing \cite{zhu2026causal}      & 87.73 & \textbf{91.37} & 88.09 & \textbf{95.95} & 86.87 & \textbf{94.75} \\
$^\dagger$Self-Forcing \cite{huang2025selfforcing} & 90.55 & \underline{65.88} & 88.55 & 29.93 & 83.70 & \underline{46.89} \\
$^\dagger$Rolling-Forcing \cite{liu2025rolling}    & 92.05 & 48.89 & 89.75 & 42.30 & 87.37 & 55.88 \\
$^\dagger$Reward-Forcing \cite{lu2025reward}       & 90.62 & 43.61 & 87.65 & \underline{44.96} & 86.17 & 52.93 \\
$^\dagger$LongLive \cite{yang2025longlive}         & 90.83 & 59.18 & 88.50 & 52.25 & 84.58 & 35.80 \\
$^\dagger$Infinite-Forcing \cite{infinite-forcing} & \underline{92.66} & 26.46 & 91.13 & 24.97 & 85.95 & 26.62 \\
$^\dagger$CausVid \cite{yin2025causvid}            & 92.51 & 40.93 & \underline{92.75} & 20.47 & 88.25 & 23.48 \\
\textbf{Steady-Forcing (Ours)}                     & \textbf{95.60} & 37.43 & \textbf{95.07} & 37.99 & \textbf{92.57} & 21.86 \\
\hline
\vspace{0em}
\end{tabular}
\caption{\textbf{Background Consistency vs.\ Dynamic Degree across extended evaluation horizons.} Causal-Forcing consistently achieves the highest Dynamic Degree alongside the lowest Background Consistency — a correlation consistent with drift-induced optical pseudo-flow rather than genuine fluid motion dynamics. Steady-Forcing achieves the highest Background Consistency at every long-horizon milestone, verifying that its lower relative Dynamic Degree reflects stable background structural anchoring rather than motion collapse artifacts. $^\dagger$~denotes baseline model weights loaded using our task-adapted fixed-camera inference protocol.}
\label{tab:supp_dd_comparison}
\end{table*}

\section{Human Evaluation Details}
\label{sec:supp_human}

\subsection{Study Protocol}
\label{sec:supp_human_protocol}

We conducted a blind four-way preference study comparing \sfm{} against
Self-Forcing~\cite{huang2025selfforcing},
Reward-Forcing~\cite{lu2025reward}, and
Rolling-Forcing~\cite{liu2025rolling}.

\textbf{Participants.}
23 participants completed all six prompts, yielding
$23 \times 6 = 138$ four-way comparisons per criterion.
The blind user study was conducted in accordance with the ethics
guidelines of the authors' institution. Participants provided
informed consent prior to participation and were free to withdraw
at any time. No personally identifiable information was collected
or retained.

\textbf{Setup.}
All four methods were anonymized (A/B/C/D) with display order randomized
independently per trial. Participants watched all four videos before selecting and could replay any video freely.

\textbf{Criteria definitions.}
\begin{itemize}
  \item \textit{Overall Quality}: best perceived video quality overall.
  \item \textit{Static-View Stability}: background does not drift,
    shift, or rotate throughout.
  \item \textit{Motion Continuity}: fluid motion persists without
    freezing or stagnating.
  \item \textit{Temporal Consistency}: no abrupt color, structure, or
    identity changes between adjacent frames.
  \item \textit{Artifact-Free Quality}: absence of repeated textures,
    carved surfaces, or other generation artifacts.
\end{itemize}

\subsection{Evaluation Prompts}
\label{sec:supp_human_prompts}

The six fixed prompts used across all participants:
\begin{enumerate}
  \item \textit{An urban storm drain canal, recorded by a static tripod camera. The heavy concrete walls and immovable steel railings remain perfectly still, while floodwater surges violently through the channel, its foamy surface reflecting the dim glow of streetlights. The camera does not move at any point. No flicker, jumps, resets, or artificial distortions appear. Temporal continuity is maintained with steady, realistic fluid dynamics throughout. [60s]}
  \item \textit{A completely fixed, tripod-mounted camera captures a flooded urban street during heavy rainfall. Rain streaks fall steadily, pooling water reflects the bridge railing and nearby vegetation. The waterline rises slowly and smoothly in one continuous upward direction, with no waves, wobble, jumps, resets, or camera movement. Concrete structures and drainage features remain static in the foreground. Urban elements such as signage and railings are visible. The atmosphere is overcast and realistic. Temporal continuity is preserved across all frames, ensuring seamless water level increase without visual artifacts. [60s]}
  \item \textit{A continuous rainy city street scene, recorded by a completely fixed, static, tripod-mounted camera. The camera is not seen, it does not move, tilt, pan, or zoom at any point. Rain falls steadily across the entire frame, creating a constant downward motion of droplets that splash gently onto the pavement. Small puddles gradually form and ripple naturally as raindrops strike their surfaces. Reflections of streetlights, urban shop signs, and passing headlights shimmer and evolve slowly on the wet ground, while all surrounding buildings, lampposts, and parked cars remain perfectly static. The atmosphere is overcast nighttime, with muted grays and soft glows of artificial light. The rain continues seamlessly without interruption, with no flicker, jumps, resets, or artificial distortions. The video maintains temporal continuity across all frames. [60s]}
  \item \textit{A cinematic wide shot of a serene forest river, its crystal-clear water flowing gently and unendingly over smooth, ancient stones. Sunlight filters through the dense canopy above, creating vibrant dancing caustics on the riverbed that shift and shimmer continuously. The surrounding mossy riverbanks and heavy gray rocks remain perfectly still and motionless, providing a sharp contrast to the unceasing downstream momentum of the water flow. The camera is completely fixed, static, and tripod-mounted. It does not move, tilt, pan, or zoom at any point. No flicker, jumps, resets, color drift, background wobble, or artificial distortions appear. The water dynamics must not stagnate, harden, or converge to a still image at any point in the sequence. Style, lighting, and scene identity remain consistent across the entire 120-second duration. Temporal continuity is preserved throughout. [120s]}
  \item \textit{A dramatic medium shot of a massive mountain waterfall, deep blue water crashing and splashing violently into a misty pool below. Thick white foam and bubbles churn and swirl rapidly in the current. The heavy, dark cliffside and the surrounding evergreen trees are completely static and immovable. The camera maintains a fixed position, capturing the unending energy and high-amplitude motion of the cascading water as it tumbles over the edge. The water dynamics must not stagnate, harden, or lose energy at any point. No background drift, color shift, style instability, or localized distortion appears. The scene remains recognizable throughout the full 120-second duration without morphing into unrelated content. Temporal continuity is preserved throughout. [120s]}
  \item \textit{A rainy urban street at night, recorded by a fixed tripod camera. The solid concrete buildings and motionless parked cars anchor the frame, while heavy rain falls endlessly, bouncing off the pavement and creating shimmering reflections. The warm glow of an orange light panel illuminates the rain-soaked street, while water streams naturally toward the gutters. The rainfall must remain continuous and physically consistent across the full 120-second generation window, without stagnation, reduction in intensity, or loss of dynamic realism. No background drift, color shift, or localized distortions may appear. [120s]}
\end{enumerate}

\subsection{Statistical Significance}
\label{sec:supp_human_stats}

Table~\ref{tab:supp_stats} reports preference rates with 95\% binomial
confidence intervals (Wilson score).
\sfm{}'s lead on all five criteria is significant ($p < 0.001$,
one-sided binomial test against chance level 0.25).


\begin{table}[htb]
\begin{center}
\small
\setlength{\tabcolsep}{4pt}
\begin{tabular}{lrrr}
\hline
Criterion & Ours & Best baseline & $p$-value \\
\hline
Overall Quality    & $0.725 \pm 0.04$ & 0.123 & ${<}0.001$ \\
Static-View Stab.  & $0.746 \pm 0.04$ & 0.101 & ${<}0.001$ \\
Motion Continuity  & $0.710 \pm 0.04$ & 0.116 & ${<}0.001$ \\
Temporal Consist.  & $0.754 \pm 0.04$ & 0.109 & ${<}0.001$ \\
Artifact-Free      & $0.710 \pm 0.04$ & 0.101 & ${<}0.001$ \\
\hline
\end{tabular}
\end{center}
\caption{\textbf{User study preference rates with 95\% CIs.}
$\pm$ values are Wilson score half-widths.}
\label{tab:supp_stats}
\end{table}


\begin{figure*}[p]
\centering

\begin{tabular}{c}
\includegraphics[width=0.85\linewidth]{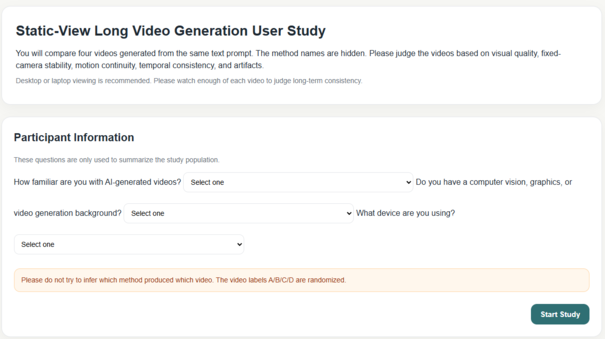} \\ [0.2em]
{\small (a) Instructions \& Criteria Definition} \\
\end{tabular}

\vspace{1.5em} 

\begin{tabular}{cc}
\includegraphics[width=0.48\linewidth]{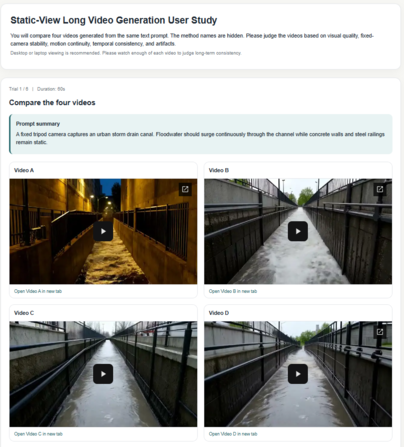} &
\includegraphics[width=0.48\linewidth]{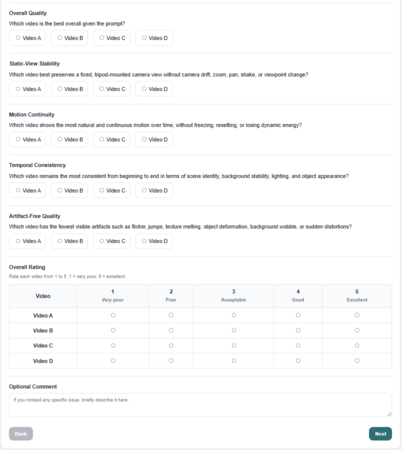} \\ [0.2em]
{\small (b) Randomized Blind Comparison} &
{\small (c) Score Assignment \& Submission} \\
\end{tabular}

\vspace{1.0em}
\caption{\textbf{Snapshots of the blind user study evaluation interface.} The study is structured across three primary stages: (a) \textbf{Instruction Phase}, detailing specific evaluation definitions (e.g., distinguishing \textit{Temporal Consistency} from \textit{Motion Continuity}); (b) \textbf{Randomized Blind Comparison}, presenting generated samples under randomized, anonymous video mapping anchors (Video A/B/C/D) to eliminate user confirmation bias; and (c) \textbf{Per-Criterion Assessment}, where participants assign 1--5 Likert ratings and cast absolute preference selections prior to sequence progression.}
\label{fig:user_study_snapshots}
\end{figure*}

\section{Ablation Details}
\label{sec:supp_ablation}

\subsection{Ablation on Teacher Model Scale up to 200s}
\label{sec:supp_ablation_teacher_scale}

\begin{figure*}[p] 
\centering
\setlength{\tabcolsep}{1pt} 
\begin{tabular}{ccc}
& {\scriptsize Steady-Forcing (Wan2.1-1.3B Teacher)} & {\scriptsize \textbf{Steady-Forcing (Wan2.1-14B Teacher) [Ours]}} \\[0em]

\parbox[c]{1cm}{\raggedleft\scriptsize $t=0\text{s}$} &
\bmvaHangBox{\includegraphics[width=0.44\linewidth]{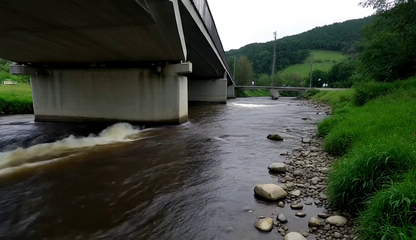}} &
\bmvaHangBox{\includegraphics[width=0.44\linewidth]{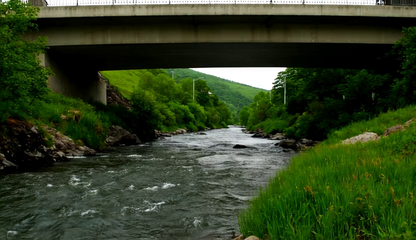}} \\ [0em]

\parbox[c]{1cm}{\raggedleft\scriptsize $t=50\text{s}$} &
\bmvaHangBox{\includegraphics[width=0.44\linewidth]{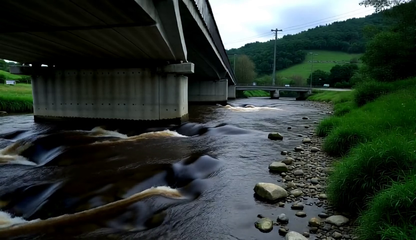}} &
\bmvaHangBox{\includegraphics[width=0.44\linewidth]{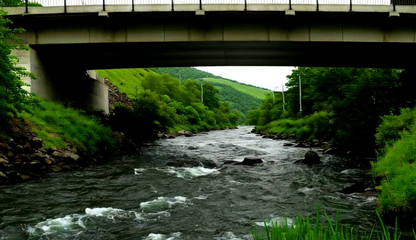}} \\ [0em]

\parbox[c]{1cm}{\raggedleft\scriptsize $t=100\text{s}$} &
\bmvaHangBox{\includegraphics[width=0.44\linewidth]{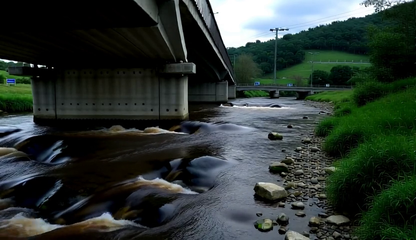}} &
\bmvaHangBox{\includegraphics[width=0.44\linewidth]{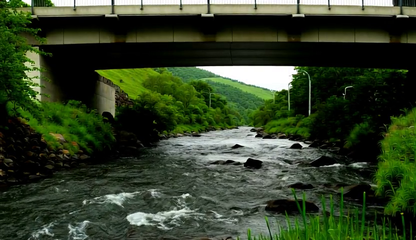}} \\ [0em]

\parbox[c]{1cm}{\raggedleft\scriptsize $t=150\text{s}$} &
\bmvaHangBox{\includegraphics[width=0.44\linewidth]{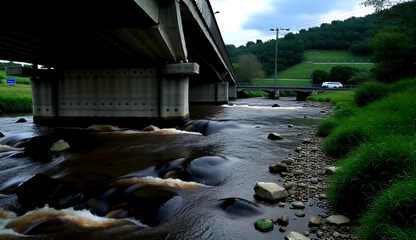}} &
\bmvaHangBox{\includegraphics[width=0.44\linewidth]{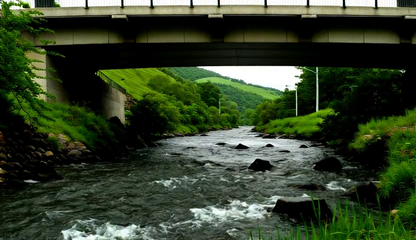}} \\ [0em]

\parbox[c]{1cm}{\raggedleft\scriptsize $t=200\text{s}$} &
\bmvaHangBox{\includegraphics[width=0.44\linewidth]{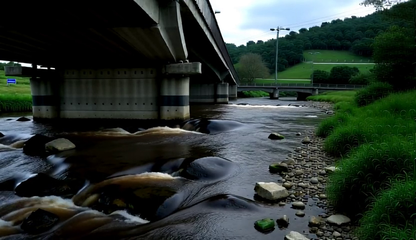}} &
\bmvaHangBox{\includegraphics[width=0.44\linewidth]{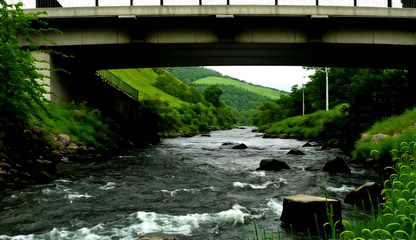}} \\

\end{tabular}
\caption{\textbf{Ablation of teacher training influence evaluated across a 200-second horizon.} Left column: video generation driven by a model distilled using our framework from the Wan2.1-1.3B (scaled-down, originally distilled from Wan2.1-14B~\cite{wan2025}) as a teacher model, leading to gradual textural \& motion stagnation over time. Right column: our final production model distilled from the original Wan2.1-14B teacher model, showing improved preservation of downstream flow details and perspective geometry across extended horizon.}
\label{fig:supp_ablation_teacher}
\end{figure*}

To analyze the impact of the distilled base capability on ultra-long horizon stability, we ablate the parameter scale of the training teacher model. Figure~\ref{fig:supp_ablation_teacher} compares generation rollouts on an identical river scene prompt evaluated out to 200 seconds. While the variant trained under the smaller Wan2.1-1.3B teacher suffers from structural collapse and loss of fluid dynamics past 100 seconds, our final model utilizing the Wan2.1-14B teacher successfully preserves rich texture boundaries and persistent downstream motion profiles.

\section{Limitation Details}

\subsection{Mid-Sequence Content Forgetting}
The stage-wise visual ablation at the 60s horizon is shown in
\textbf{Main Fig.~6}. Mid-sequence content forgetting arises because
the fixed-size memory (V-Sink + EMA-Sink + local window) retains the
initial frame and recent context but has no explicit mechanism to
recall content generated in between. Frames from the middle of a long
sequence are progressively compressed into the EMA-Sink via
exponential moving average and eventually discarded at each KV Flush.
This is an inherent trade-off of the bounded-memory design: it enables
$\mathcal{O}(w+s)$ constant memory but cannot preserve fine-grained
scene details across the full generation duration.

\section{Method Details}

\subsection{DMD Training Objective}
\label{sec:supp_dmd}

The Distribution Matching Distillation loss used in
Algorithm~\ref{alg:training} aligns the student rollout distribution
$p_\theta$ with the teacher score function at a matched diffusion noise
level $t$:
\begin{equation}
\label{eq:supp_dmd}
  \mathcal{L}_{\text{DMD}} = \mathbb{E}_{t,\,x\sim p_\theta}
  \Bigl[
    \bigl(\mathbf{s}_\phi(x_t) - \mathbf{s}_\psi(x_t)\bigr)
    \cdot \nabla_\theta \log p_\theta(x)
  \Bigr]
\end{equation}
where $\mathbf{s}_\phi$ is the frozen teacher score function (Wan2.1-14B),
$\mathbf{s}_\psi$ is a separately maintained fake score network updated
adversarially, and $x \sim p_\theta$ are student-generated sequences from
the Self-Forcing unroll with the full inference memory configuration active
(V-Sink, EMA-Sink, local window). The objective is supervised at the
sequence level rather than frame-by-frame, aligning the full rollout
distribution rather than individual frames in isolation.
The fake score network $\mathbf{s}_\psi$ is updated to distinguish
student outputs from teacher-scored samples; the generator $G_\theta$
is updated to minimize the score gap. Generator and critic learning
rates follow the values in Table~\ref{tab:supp_arch}.

\subsection{EMA-Sink Sliding Attention Window}
The local attention window $w$ is set to 21 latent frames (one
complete 5-second clip, matching the Wan2.1~\cite{wan2025} training horizon)
throughout both training and inference. As frames are evicted beyond this 21-frame window, their key-value pairs are fused into the EMA-Sink's fixed-size
compressed state (2 latent frames of storage) via exponential
moving average, maintaining a running kinetic summary without
growing memory cost. Following~\cite{lu2025reward}, the EMA update is applied per attention layer and shared across heads within each layer.

\subsection{Inference Pseudocode}
\label{sec:supp_pseudo}

Algorithms~\ref{alg:inference} and~\ref{alg:training} present the
complete \sfm{} inference loop and distillation procedure.
Equation numbers refer to the main paper.

\begin{algorithm2e}[htb]
\DontPrintSemicolon
\SetKwInOut{Input}{Input}
\Input{Prompt $c$; model $G_\theta$; total frames $T$; block size $\Delta$;
       $w{=}21$; $\alpha{=}0.99$; $N_{purify}{=}21$; $m{=}5$}
\BlankLine
Generate Frame~0: $x_0 \leftarrow G_\theta(c)$\;
Initialize V-Sink: $S_\text{fixed} \leftarrow \mathrm{KV}(x_0)$\;
Initialize EMA-Sink: $S_0^K \leftarrow \mathbf{0}$;\; $S_0^V \leftarrow \mathbf{0}$\;
KV cache: $\mathcal{C} \leftarrow [S_\text{fixed}]$\;
\BlankLine
\For{$i = 1, 2, \ldots, T/\Delta$}{
  \tcc{Periodic KV Flush (Eqs.~3--4)}
  \If{$i \bmod N_{purify} = 0$}{
    $\mathcal{C} \leftarrow [S_\text{fixed} \;;\; \mathrm{KV}(x_{i-m:i})]$\;
    $S_i^K \leftarrow \mathbf{0}$;\; $S_i^V \leftarrow \mathbf{0}$\;
  }
  \BlankLine
  \tcc{Build global context (Eq.~2)}
  $K^\mathrm{g} \leftarrow [S_\text{fixed}^K \;;\; S_i^K \;;\; K_{i-w+1:i}]$\;
  $V^\mathrm{g} \leftarrow [S_\text{fixed}^V \;;\; S_i^V \;;\; V_{i-w+1:i}]$\;
  \BlankLine
  \tcc{Generate next chunk}
  $x_{i+1:i+\Delta} \leftarrow G_\theta(c,\,K^\mathrm{g},\,V^\mathrm{g})$\;
  \BlankLine
  \tcc{EMA-Sink update (Eq.~1)}
  $K_e, V_e \leftarrow \mathrm{KV}(x_{i-w})$\;
  $S_{i+1}^K \leftarrow \alpha S_i^K + (1{-}\alpha)K_e$\;
  $S_{i+1}^V \leftarrow \alpha S_i^V + (1{-}\alpha)V_e$\;
  $\mathcal{C} \leftarrow \mathcal{C} \cup \mathrm{KV}(x_{i+1:i+\Delta})$\;
}
\Return $\{x_0, x_1, \ldots, x_T\}$\;
\caption{\sfm{} Autoregressive Inference}
\label{alg:inference}
\end{algorithm2e}

\begin{algorithm2e}[t]
\DontPrintSemicolon
\SetKwInOut{Input}{Input}
\Input{Student $G_\theta$; frozen teacher $G_\phi$; prompt set
       $\mathcal{P}$; iterations $N{=}6000$; neg.\ prompt $c^-$}
\BlankLine
Initialize $G_\theta$ from Reward-Forcing~\cite{lu2025reward} checkpoint\;
\BlankLine
\For{step $= 1 \ldots N$}{
  $c \sim \mathcal{P}$\;
  \tcc{Self-Forcing unroll with full inference config active}
  $\mathbf{x} \leftarrow \mathrm{SteadyForcingInference}(c, G_\theta)$\;
  \BlankLine
  \tcc{DMD loss (Eq.~\ref{eq:supp_dmd})}
  $\hat{s}_\phi \leftarrow G_\phi(\mathbf{x}_{t_k})$\;
  $\hat{s}_\psi \leftarrow G_\psi(\mathbf{x}_{t_k})$\;
  $\mathcal{L} \leftarrow \mathbb{E}\bigl[(\hat{s}_\phi - \hat{s}_\psi)
    \cdot \nabla_\theta \log p_\theta(\mathbf{x})\bigr]$\;
  \BlankLine
  Update $\theta \leftarrow \theta - \eta_g \nabla_\theta \mathcal{L}$\;
  Update $\psi \leftarrow \psi - \eta_c \nabla_\psi \mathcal{L}_\mathrm{fake}$\;
}
\caption{\sfm{} Rewarded DMD Distillation}
\label{alg:training}
\end{algorithm2e}

\section{Implementation and Training Details}
\label{sec:supp_impl}

\subsection{Full Architecture Specification}
\label{sec:supp_impl_arch}

\begin{table}[htb]
\begin{center}
\small
\setlength{\tabcolsep}{5pt}
\begin{tabular}{ll}
\hline
Component & Specification \\ \hline
Teacher model              & Wan2.1-T2V-14B~\cite{wan2025} \\
Student backbone           & Wan2.1-T2V-1.3B~\cite{wan2025} \\
Student init.              & Reward-Forcing~\cite{lu2025reward} \\
Temporal compression       & $4\times$ \\
Spatial compression        & $8\times$ \\
Max RoPE indices           & 1024 per dimension \\
Frames per block ($\Delta$)& 3 latent frames \\
Number of training frames  & 21 \\
Sink size                  & 3 \\
Local attention size       & 21 \\
EMA decay ($\alpha$)       & 0.99 (per layer, shared across heads) \\
Flush period ($N_{purify}$)& 21 blocks \\
Motion anchor ($m$)        & 5 frames \\
Training iterations        & 6{,}000 \\
Batch size                 & 8 (8$\times$1) \\
Denoising steps            & 4 (uniform schedule) \\
Generator LR               & $2.0\times10^{-6}$ \\
Critic LR                  & $4.0\times10^{-7}$ \\
Optimizer                  & AdamW \\
Hardware                   & $8\times$ NVIDIA A100 (80\,GB) \\
Training time              & $\approx$67 hours \\ \hline
\end{tabular}
\end{center}
\caption{\textbf{Full architecture and training configuration.}}
\label{tab:supp_arch}
\end{table}

\subsection{Data-Free Prompt Corpus Structure}
\label{sec:supp_impl_corpus}

The 21{,}000-prompt training corpus is constructed combinatorially from
four semantic pools:

\begin{itemize}
  \item \textbf{Static anchors} (${\approx}50$ entries): mountains,
    canyon walls, rocky coastlines, dense forest, ancient stone
    bridges, volcanic plateaus, and similar.
  \item \textbf{Fluid flows} (${\approx}40$ entries): river currents,
    waterfalls, ocean waves, rain, lava streams, smoke columns,
    and similar.
  \item \textbf{Atmospheric conditions} (${\approx}30$ entries):
    golden hour, overcast, misty morning, moonlit, stormy, and
    similar.
  \item \textbf{Global locations} (${\approx}35$ entries):
    Norwegian fjords, Himalayan foothills, Amazon basin,
    Icelandic highlands, Asian city, and similar.
\end{itemize}

\noindent
Every prompt follows the structure:
\begin{center}
    \texttt{[Static anchor] + [Fluid flow] + [Atmospheric condition] +} \\
    \texttt{[Location] + [Fixed-camera constraint]}
\end{center}

The fixed-camera constraint appended to every prompt:
\textit{``Completely stationary, tripod-mounted camera.
No camera movement, no panning, no tilting, no zooming.''}

The negative prompts used during both distillation and inference:
\textit{``repetitive round textures, ground artifacts, motion
stagnation, frozen motion, static water, hardened flow, camera
movement, panning, tilting, zooming, shaky footage, blurry,
low quality.''}

\clearpage
\section{Ethical Considerations}
\label{sec:supp_ethics}

\textbf{Synthetic ambient media.}
\sfm{} generates photorealistic long-form nature video from a stationary
viewpoint. Potential misuse includes fake live-stream content or synthetic
surveillance footage. Any deployment should disclose that content is
AI-generated.

\textbf{Environmental representation.}
The model inherits the geographic biases of Wan2.1's training distribution;
nature scenes from underrepresented regions may be rendered with lower
fidelity.

\textbf{Computational cost.}
Training required approximately 67 wall-clock hours on $8\times$ A100 (80\,GB) GPUs.

\textbf{Scope.}
This work targets passive ambient generation and is not designed for
interactive or action-conditioned world simulation.


\section{Comparison with Concurrent Work: Grounded Forcing}
\label{sec:supp_grounded}

Grounded Forcing~\cite{chen2026grounded} is a concurrent work
that shares the same Wan2.1-T2V-1.3B backbone~\cite{wan2025} and uses a dual-memory
KV cache with a fixed-RoPE-index global anchor, making it architecturally
adjacent to Steady-Forcing. We clarify the relationship here.

\textbf{Different goals.} Grounded Forcing targets semantic and identity
consistency in interactive, multi-shot narrative generation — maintaining
character identity (e.g., a specific person or object) across prompt
switches and scene transitions. Steady-Forcing targets the
stability--motion trade-off in passive fixed-camera nature streams,
where the scene never legitimately changes and the challenge is
preserving static background layout while sustaining fluid motion.

\textbf{Different global memory design.} Grounded Forcing's Global
Consistency Memory (GCM) is \emph{dynamically updated} when newly
generated frames have high semantic novelty relative to existing anchors
(diversity-aware replacement). This allows the memory to evolve as new
characters or scenes are introduced. Steady-Forcing's V-Sink is
\emph{permanently immutable}: Frame~0 is retained without modification
for the full rollout duration, which is the correct design for a
fixed-camera passive stream where any change to the global anchor would
permit background drift.

\textbf{Different evicted-frame handling.} Grounded Forcing's Local
Temporal Memory (LTM) is a standard sliding window — evicted frames
are discarded. Steady-Forcing's EMA-Sink compresses all evicted frames
into a continuously updated global kinetic summary via exponential
moving average, preserving fluid momentum from content that has left
the local window.

\textbf{Different cache management.} Grounded Forcing's Asymmetric
Proximity Recache (APR) is designed for smooth semantic inheritance
during prompt transitions. Steady-Forcing's Periodic KV Flush is a
scheduled error-purging strategy to prevent accumulated cache
contamination from hardening into repeated texture artifacts — a
failure mode that does not arise in Grounded Forcing's interactive
use case.

\textbf{Summary.} The architectural similarity (dual-memory + fixed-index
anchor) reflects convergent design reasoning applied to different
problems. The mechanisms and their rationales differ substantially,
and the two methods are complementary rather than competing.


\end{document}